\documentclass[letterpaper, 10 pt, conference]{ieeeconf}

\newcommand{\Logmap}{\operatorname{Logmap}}
\IEEEoverridecommandlockouts
\usepackage[table, dvipsnames]{xcolor}
\usepackage{amsmath}
\usepackage{graphicx}
\usepackage{adjustbox}
\usepackage{multicol}
\usepackage{tabularx}
\usepackage{caption}
\usepackage{amssymb}
\usepackage{cellspace}
\usepackage{hyperref}
\usepackage{cite}
\usepackage{multirow}
\usepackage{float}
\usepackage{subcaption}
\usepackage{soul}
\let\labelindent\relax
\usepackage{enumitem}
\usepackage{xspace}
\usepackage{ulem}
\usepackage[ruled,vlined]{algorithm2e}
\usepackage{bm}

\usepackage{booktabs}
\usepackage[table]{xcolor}
\definecolor{best}{RGB}{158,214,156}     % light green
\definecolor{second}{RGB}{253,230,201}   % light orange
\definecolor{sectiongray}{gray}{0.94}

\newcommand{\coolname}{Steer2Grasp\xspace}

\title{\LARGE \bf
\coolname: Inference-Time Embodiment-Aware Steering for Diverse Physically Feasible Grasp Diffusion
}

\author{
\shortstack[c]{
Vignesh Vembar$^{1*}$, Ayush Kaura$^{1*}$, A Padmaprabhan$^{1*}$, Siddharth Sinha$^{1}$, \\
Kailash Nagarajan$^{1}$, Keshab Patra$^{1}$, Md Faizal Karim$^{2}$, K Madhava Krishna$^{1}$
}
\thanks{$^*$ Equal Contribution}
\thanks{$^{1}$ Robotics Research Center, IIIT Hyderabad}
\thanks{$^{2}$ Johns Hopkins University}
}

\begin{document}

\maketitle
\thispagestyle{empty}
\pagestyle{empty}

%%%%%%%%%%%%%%%%%%%%%%%%%%%%%%%%%%%%%%%%%%%%%%%%%%%%%%%%%%%%%%%%%%%%%%%%%%%%%%%%

\begin{abstract}
Current grasp diffusion models provide rich priors for generation, yet their object-centric approach can violate the kinematic and collision constraints imposed by the embodiment and the environment. Existing embodiment-aware methods primarily perform local corrections around generated grasps through gradient guidance or optimization, making it difficult to recover from fundamentally infeasible modes. We present Steer2Grasp, a training-free, embodiment-agnostic framework for inference-time grasp steering that adapts a frozen Cartesian grasp diffusion model using deployment-specific rewards. Through Feynman-Kac (FK) inspired particle reweighting and resampling, the method reallocates population mass from infeasible to high-reward grasp modes, enabling population-level mode transitions without modifying the pretrained diffusion model or requiring differentiable constraints.
The framework enables a unified treatment for single and dual arm grasping through reachability and collision aware rewards, followed by gradient free gripper level local refinement. Across diverse objects, robot embodiments, and constrained environments, our method substantially improves feasible grasp generation while maintaining proximity to the underlying grasp prior.  Project Page: \href{https://steer2grasp.github.io}{steer2grasp.github.io} 
\end{abstract}

%%%%%%%%%%%%%%%%%%%%%%%%%%%%%%%%%%%%%%%%%%%%%%%%%%%%%%%%%%%%%%%%%%%%%%%%%%%%%%%%
\section{Introduction}
\label{sec:intro}
Manipulation of everyday objects requires not only identifying stable grasp configurations, but ensuring that they are executable by the robot(s). A grasp is \textit{feasible} if it is force-closure stable, admits a valid robot joint configuration, and is collision-free with respect to the environment.
With regard to the first criterion, early approaches demonstrated the effectiveness of analytic grasp metrics and learned grasp quality models \cite{grasping_review}, while more recent diffusion-based methods learn multimodal distributions over 6-DoF grasp poses \cite{se3diff,graspgen}. This paradigm has been extended to large-object and bimanual manipulation, with newer datasets \cite{da2,dg16m} enabling generation of coordinated grasp pairs using diffusion models such as DAGDiff \cite{dagdiff} and transformer networks such as BiGraspFormer \cite{bigraspformer}.

\begin{figure}[!t]
    \centering
    \includegraphics[width=0.9\columnwidth]{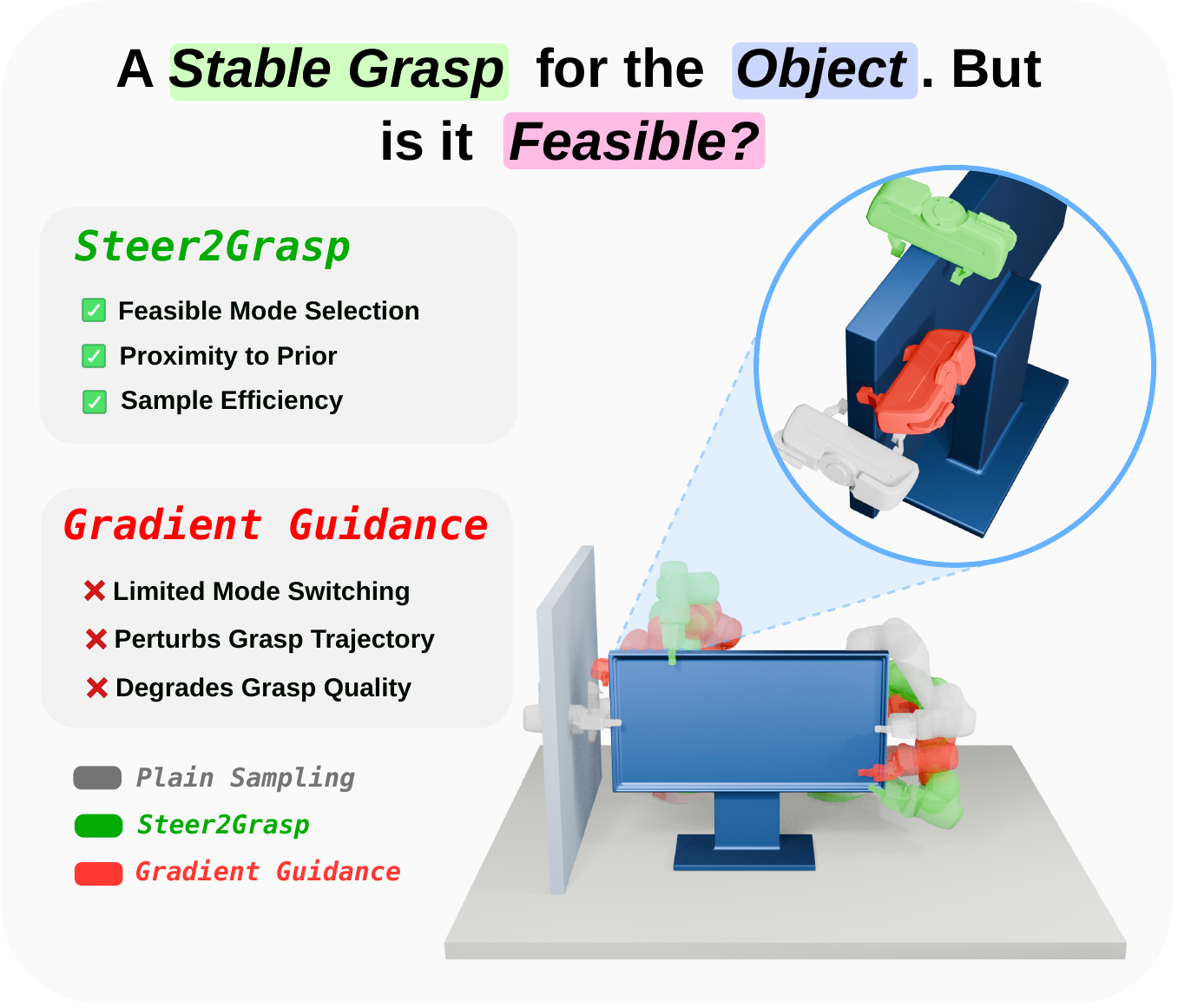}
    \captionsetup{font=footnotesize}
    \caption{Object-centric grasp models may generate force closure valid grasps that are unreachable or colliding for a specific robot embodiment configuration and environment. Local gradient guidance addresses this by perturbing individual denoising trajectories toward feasibility, but such local updates can degrade grasp quality and struggle to move between distant grasp modes.
    \textbf{\coolname} instead reweights and resamples the particle population, reallocating probability mass toward feasible modes while leaving the pretrained diffusion model unchanged.}
    \label{fig:teaser}
    \vspace{-20pt}
\end{figure}

Despite this progress in generating stable and diverse grasps, most grasp synthesis methods remain largely \textit{arm agnostic}: they operate in an object-centric representation and do not explicitly account for robot kinematics or deployment-time environmental constraints. As a result, a grasp that is geometrically stable in task space, may be unreachable by the robot, or may require a robot joint configuration that collides with surrounding obstacles. This problem is substantially harder in the bimanual setting, where feasibility is determined by the joint configurations of both arms, and grasps that are individually reachable and collision free may still violate inter arm or coordination constraints, making the jointly feasible set a small, sparse subset of full grasp space.

Three natural strategies have emerged for closing this gap, each with a fundamental limitation. The first is to \underline{incorporate embodiment directly into training} by filtering grasp data for a specific robot and scene prior to training \cite{reachable_classifier}, allowing the model to learn a feasibility-aware distribution from the outset. While effective, this ties the learned distribution to a particular embodiment, and deploying on a different robot or in a different scene requires collecting new data and retraining from scratch. The second is \underline{rejection filtering}, leaving the prior unchanged and discarding generated grasps that violate kinematic or collision constraints. This avoids retraining but is inherently sample-inefficient, since the feasible set can constitute a small, disconnected subset of the full grasp space. Finally, recent \underline{inference time diffusion guidance} methods \cite{arm_aware,embodisteer,ggp} have shown that the iterative denoising process can be guided by task-specific objectives, enabling generated grasps to satisfy constraints without retraining the underlying prior. However, such guidance comes at the cost of modifying the learned grasp distribution: aggressive guidance can pull samples away from high-probability regions, potentially producing less stable or lower-quality grasps. Moreover, gradient guidance is fundamentally local: it can refine a sample toward a nearby feasible region, but cannot transport it across an infeasible region to reach a distant feasible mode, a critical limitation when feasible grasps are sparse or disconnected, as is common for geometrically constrained objects and in bimanual grasping as shown in Fig.~\ref{fig:teaser}.

In this work, we introduce \textbf{\coolname}, an inference-time steering framework for zero-shot embodiment-aware grasp generation applicable to any grasp diffusion network. Rather than modifying the learned prior or post-hoc filtering, feasibility constraints are expressed as \textit{potential functions} evaluated at intermediate denoising steps, which softly reweight the particle population and trigger resampling to concentrate probability mass on particles likely to yield feasible grasps. Unlike gradient guidance ~\cite{arm_aware}, which directly perturbs the denoising trajectory, embodiment constraints govern which particles survive and propagate, leaving individual denoising trajectories faithful to the learned prior. 
This also enables population-level mode reallocation in a way that local gradient steps cannot achieve, particles tracking infeasible modes are removed and replaced by resampling from survivors.
To further reduce gripper collisions which are inherently local, we employ a refinement stage in the last stages of the steering that correct fine grained collisions.
We validate \coolname with refinement on GraspGen \cite{graspgen} for single-arm grasping and
DAGDiff \cite{dagdiff} for bimanual grasping, improving feasibility over the
base model by $\sim$32pp and $\sim$43pp, and over gradient-based guidance by $\sim$22pp and $\sim$35pp
respectively.

To summarize our contributions:
\begin{enumerate}
    \item An inference-time steering framework that converts object-centric diffusion grasps into embodiment-aware ones, improving over baselines in both single- and dual-arm grasping. We are the first to extend embodiment-aware steering to dual-arm grasps, where feasibility must hold jointly across both arms.
    \item A two-level steering mechanism that combines population-level mode reallocation, enabling transitions beyond local gradient corrections, with a final refinement stage that resolves fine-grained collisions.
    \item Extensive simulation on the Franka Panda and UR5e across two priors
    (GraspGen and DAGDiff) and diverse task-space settings, with real-world
    demonstrations on the xArm7, showing our framework's robustness.
\end{enumerate}

\section{Related Work}

\subsection{Inference Time Alignment in Diffusion Models}
Diffusion models provide flexible generative priors adaptable to downstream objectives during inference without modifying their learned parameters. 
Classifier and classifier-free guidance \cite{dhariwal_classifier_guidance,cfg}
steer samples toward target conditions but require a differentiable classifier
or conditions fixed at training time. Reward-guided methods
\cite{janner2022diffuser,uehara_reward} extend this to arbitrary task rewards,
but still guide with gradients of a learned reward or value model, while
FK-Steering \cite{fksteer_singhal} removes the gradient requirement entirely via
Feynman--Kac particle resampling with non-differentiable rewards. We leverage this paradigm for grasp generation by introducing deployment specific rewards induced by robot embodiment and environmental constraints. 

\subsection{Cross-Embodiment Generalization for Grasping}
One aspect of cross-embodiment grasping is generalization across gripper embodiments~\cite{graspgenx}, while arm-level kinematic and environmental feasibility are typically handled separately.~\cite{se3diff} integrates learned grasp costs with collision and joint constraints for joint grasp and motion optimization, while projection-based approaches enforce feasibility by projecting samples onto constraint-satisfying regions during denoising~\cite{dpcc}. Gradient-based approaches use reachability and arm--environment constraints to guide a pretrained arm-agnostic diffusion model toward feasible grasps \cite{arm_aware}. \cite{ggp} further formulates the inference-time diffusion guidance as a constrained non-linear optimization problem, replacing reverse-step perturbations with optimized corrections that enforce embodiment and task specific feasibility. However, these approaches primarily refine individual grasps toward local feasibility, which can fail when feasible grasps lie in a different mode of the grasp distribution. Our work instead steers a population of grasp candidates 
using embodiment and environment feedback, enabling transitions between feasible modes while maintaining closeness to the pretrained grasp prior.

\section{Preliminaries}
\label{sec:preliminaries}

We build on two grasp diffusion models as our base priors. For single-arm grasping, we use GraspGen \cite{graspgen}, which generates 6-DoF grasp poses $H\in SE(3)$ conditioned on an object pointcloud $P$ via DDPM \cite{ddpm} in $T=10$ steps. For dual-arm grasping, we use DAGDiff \cite{dagdiff}, which jointly generates grasp pairs $H=(H_1,H_2)\in SE(3)\times SE(3)$ via score-based Langevin dynamics in $T=225$ steps. In both cases, an estimate of the clean grasp at each denoising step is obtained via Tweedie's formula~\cite{tweedie, universal_guidance}. For GraspGen and DAGDiff, respectively,
\begin{subequations}
\label{eq:tweedie}
\begin{align}
\hat{H}_0(H_t)
&=
\frac{1}{\sqrt{\bar{\alpha}_t}}
\left(
H_t -
\sqrt{1-\bar{\alpha}_t}\,
\epsilon_\theta(H_t,P,t)
\right)
\label{eq:ddpm_tweedie}
\\
\hat{H}_0(H_t)
&=
\mathrm{Expmap}_2\!\left(
\mathrm{Logmap}_2(H_t)
+
\sigma_t^2\,
s_\theta(H_t,P,t)
\right)
\label{eq:tweedie_score}
\end{align}
\end{subequations}
where $t$ is the diffusion step, $\bar{\alpha}_t$ and $\sigma_t$ are the respective noise schedules, $\epsilon_\theta$ and $s_\theta$ denote the GraspGen noise predictor and DAGDiff score function, and $\mathrm{Logmap}_2$ and $\mathrm{Expmap}_2$ operate component-wise on $SE(3)\times SE(3)$. The Tweedie estimates $\hat{H}_0(H_t)$ play a central role in our method, as they provide a clean grasp prediction at each intermediate step against which feasibility rewards can be evaluated. 

\section{Method}
\label{sec:method}
\begin{figure*}[t!]
    \centering
    \includegraphics[width=\textwidth]{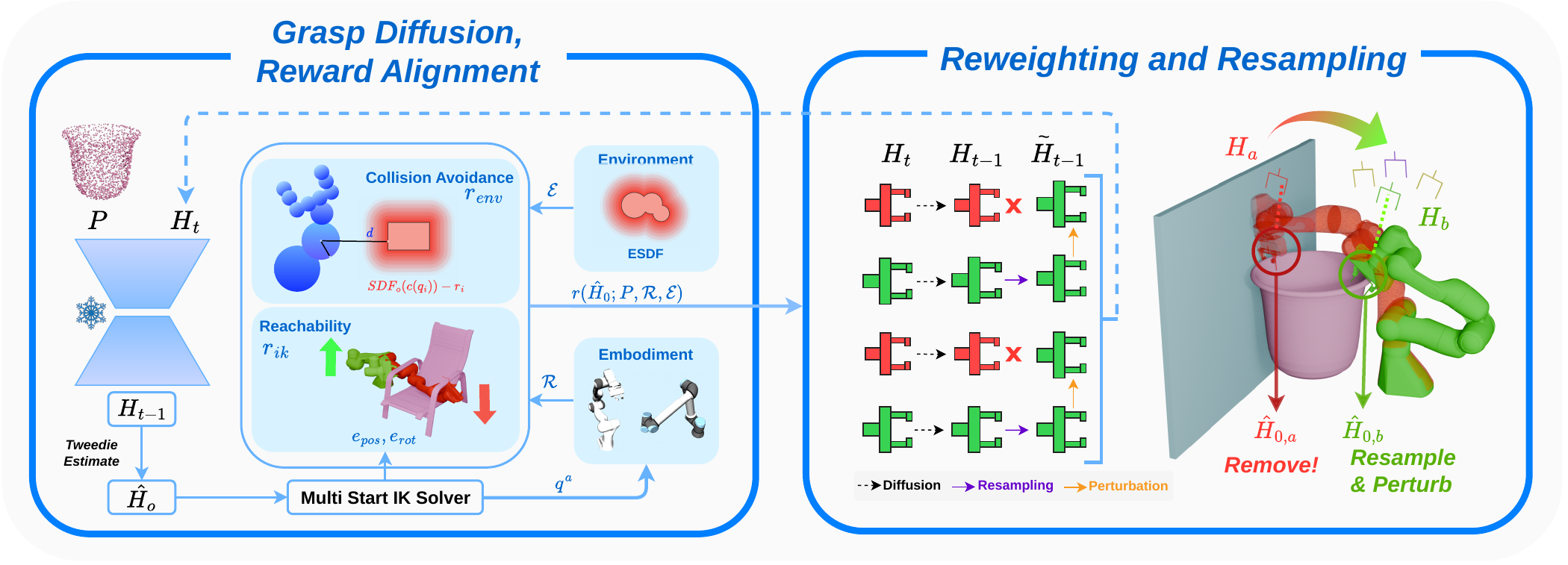}
    \captionsetup{font=footnotesize}
    \caption{\textbf{Overview of the proposed method:} Given an full object point cloud $P$ and a pretrained Cartesian space grasp diffusion model, the reverse diffusion produces grasp hypotheses $H_{t-1}$ which are mapped to their clean grasp estimates $\hat{H}_0(H_t)$ using the Tweedie estimate, which is evaluated against deployment specific feasibility rewards. The resulting reward $\mathbf{r}(\hat{H}_0;P,\mathcal{R},\mathcal{E})$ combine embodiment and environment information through multi-start IK, reachability and signed geometric clearances. These rewards instantiate the Feynman--Kac twisting potential used to re-weight and resample ($\mathbf{H}_{b}$) the particle population in turn removing particles ($\mathbf{H}_{a}$). Thus embodiment and environment constraints act through population level reweighting and resampling rather than modifying individual denoising transitions, enabling mode level steering toward feasible regions while retaining the multi modal structure of the pretrained grasp prior. }
    \label{fig:method_overview}
    \vspace{-20pt}
\end{figure*}
\label{subsec:problem_setup}
Given a point-cloud observation $P \in \mathbb{R}^{N_{p} \times3}$ of an object, a robot embodiment $\mathcal{R}$, and a deployment environment $\mathcal{E}$, our goal is to generate a grasp (or a coordinated grasp pair) that remains close to the pretrained task-space grasp prior $p^{\theta}_{0:T}(H_{0:T}\mid P)$, while satisfying the environment and embodiment specific constraints, where $H_t$ denotes the grasp state at diffusion timestep $t$, and $T$ is the total number of diffusion steps. 
Importantly, the prior is independent of the robot embodiment and the deployment environment; these quantities are introduced only at inference time to evaluate and steer grasps toward executable solutions. We assume access to the robot's forward and inverse kinematics, allowing a Cartesian wrist pose $H^a\in SE(3)$ to be mapped to a feasible joint configuration $q^a\in\mathcal{Q}^a$ for each arm $a\in\{L,R\}$, where $\mathcal{Q}^a$ is the joint limit constrained configuration space of arm $a$.

Following the reward-tilted diffusion sampling through Twisted Sequential Monte Carlo \cite{wu_tds_smc_23}, we formulate the deployment time, embodiment aware and physically feasible grasp generation as sampling from a feasibility tilted version of the pretrained grasp prior. We seek to align the prior $p^\theta$ at inference time using a deployment reward $\mathbf{r}(H_0;P,\mathcal{R},\mathcal{E})$, which evaluates the feasibility of a grasp for the target embodiment and the environment and seek a reward-aligned distribution by regularizing the deployment reward against the pretrained grasp prior,
\begin{equation}
p^*
=
\arg\max_p
\left[
\mathbb{E}_{p}[\mathbf{r}(H_0;P,\mathcal{R,}\mathcal{E})]
-
\tau D_{\mathrm{KL}}
\left(
p\,\|\,p^\theta
\right)
\right]
\label{eq:kl_objective}
\end{equation}
where $\tau>0$ controls the strength of the reward alignment relative to adherence to the prior. The optimal solution $p^*=p^*_{0:T}(H_{0:T}\mid P,\mathcal{R},\mathcal{E})$ is an exponential tilt of the pretrained path distribution~\cite{fksteer_singhal},
\begin{equation}
p^*
\propto
p^\theta
\exp\!\left(
\mathbf{r}(H_0;P,\mathcal{R},\mathcal{E})/\tau
\right).
\label{eq:reward_tilt}
\end{equation}
The corresponding diffusion-time twisting potential is obtained by
conditioning this terminal tilt on the intermediate state $H_t$. Following \cite{fksteer_singhal,cdm_smc}, the corresponding twisting potential at
diffusion timestep $t$ is
\begin{equation}
\psi_t^*(H_t)
=
\mathbb{E}_{p^\theta(H_0\mid H_t,P)}
\left[
\exp\!\left(
\frac{\mathbf{r}(H_0;P,\mathcal{R},\mathcal{E})}{\tau}
\right)
\right]
\label{eq:optimal_twist}
\end{equation}

Since this conditional expectation is intractable, we approximate the terminal grasp using its Tweedie estimate and realize the resulting reward aligned distribution through inference time steering. The following sections describe the inference time reward alignment procedure \ref{sec:Inference time Reward Alignment}, the local refinement procedure \ref{sec:local_refinement} and the deployment specific feasibility rewards \ref{subsec:feasibility_rewards}.
\subsection{Inference Time Reward Alignment}\label{sec:Inference time Reward Alignment}
The objective in Eq .~\eqref{eq:kl_objective} defines the desired reward-aligned distribution, but direct sampling from this distribution is intractable. We therefore approximate its diffusion time twisting potential and realize the resulting distribution through inference time reward alignment. For continuous diffusion, evaluating the conditional expectation in Eq.~\eqref{eq:optimal_twist} can be expensive. We approximate the twist using the denoised final grasp estimate obtained from the Tweedie's formula \cite{tweedie} from Eq.~\eqref{eq:ddpm_tweedie}, and ~\eqref{eq:tweedie_score}, giving a tractable approximation 
\begin{equation}
    \psi_t(H_t;P,\mathcal{R},\mathcal{E})
    \approx
    \exp\left(\mathbf{r}(\hat{H}_0(H_t);P,\mathcal{R},\mathcal{E})/\tau
    \right)
    \label{eq:twist_tweedie}
\end{equation}
to the reward-aligned intermediate distribution. We next instantiate the reward $\mathbf{r}(\hat{H_0}(H_t);P,\mathcal{R},\mathcal{E})$ using robot and environment specific feasibility rewards evaluated on the denoised estimate $\hat{H_0}$. The deployment reward combines kinematic reachability with environment, and object collision related geometric terms. These rewards are evaluated on the denoised grasp estimate and used to construct the FK potential for particle reweighting and resampling. The specific reward terms, their scaling, and aggregation for single- and dual-arm grasps are detailed in
Sec.~\ref{subsec:feasibility_rewards}.

Given the deployment reward $\mathbf{r}(\hat{H}_0;P,\mathcal{R},\mathcal{E})$, we approximate sampling from the corresponding reward-aligned distribution without modifying the pretrained diffusion model. At diffusion timestep $t$, we maintain a population of $N$ particles, $\mathcal{H}_t=\left\{H_t^{(n)}\right\}_{n=1}^{N}$ representing multiple hypotheses under the pretrained grasp distribution. Each particle is first propagated using the pretrained reverse diffusion step,
\begin{equation}
    H_{t-1}^{(n)}
    \sim
    p^{\theta}
    \left(
        H_{t-1}\mid H_t^{(n)},P,t
    \right)
\end{equation}
To align this population toward deployment-feasible configurations, we
evaluate the twisting potential $\psi_{t-1}\left(
        H_{t-1}^{(n)};P,\mathcal{R},\mathcal{E}
    \right)$  for each particle using the Tweedie-based
approximation in Eq.~\eqref{eq:twist_tweedie}. We use the ratio of the current and previous steering event twisting potentials to obtain the incremental difference potential
\begin{equation}
    \phi_{t-1}^{(n)}
    =
        \psi_{t-1}^{(n)}
    /
        \psi_{\mathrm{prev}}^{(n)}
    =
    \exp\left(
        \left(
            \mathbf{r}_{t-1}^{(n)}
            -
            \mathbf{r}_{\mathrm{prev}}^{(n)}
        \right)/\tau
    \right)
    \label{eq:difference_potential}
\end{equation}
where $\mathbf{r}_{prev}^{(n)}$ is the reward associated with the particle $n$ from the previous steering event (0 for first event). This incremental potential assigns larger weights to particles whose deployment reward improves between successive steering events. The propagated particles are assigned normalized importance weights according to
\begin{equation}
    w_{t-1}^{(n)} =
        \phi_{t-1}^{(n)}/\sum_{j=1}^{N}\phi_{t-1}^{(j)}
    \label{eq:particle_weights}
\end{equation}
and the particle population is resampled according to $\{w_{t-1}^{(n)}\}_{n=1}^{N}$. The reweight and resampling step is applied intermittently at a predefined set of diffusion time steps $\mathcal{T}_s \subseteq \{1,\ldots T\}$. At $t \notin \mathcal{T}_s$, particles follow the unmodified reverse diffusion processes. Resampling is performed with replacement, allowing high reward particles to produce multiple offspring. To mitigate the resulting particle impoverishment caused by resampling, duplicated offspring are perturbed in the tangent space $\mathfrak{se}(3)$ as
\begin{equation}
    \tilde{H}_{t-1}^{(n)}
    =
    \operatorname{Expmap}
    \left(
        \operatorname{Logmap}\!\left(
            \bar{H}_{t-1}^{(n)}
        \right)
        +
        \gamma \sigma_t \epsilon
    \right),
    \label{eq:particle_rejuvenation}
\end{equation}
where $\bar{H}^{(n)}_{t-1}$ is the resampled particle, $\epsilon\sim\mathcal{N}(0,I_6)$, and $\sigma_t$ is the current diffusion noise scale, and $\gamma$ controls the perturbation magnitude. For dual arm grasps, the same operation is applied jointly on $SE(3)\times SE(3)$ using $\mathrm{Logmap_2,Expmap_2}$, and $\epsilon \sim \mathcal{N}(0,I_{12})$. The intermittent steering schedule $\mathcal{T}_s$ allows particles to evolve under the pretrained reverse diffusion between interventions, retaining proximity to the learned multimodal prior, while reward-based alignment reallocates the particles toward higher feasibility modes and tangent-space perturbations maintain local diversity around duplicated particles. Figure~\ref{fig:method_overview} summarizes this inference-time process,
showing how the pretrained reverse diffusion trajectory is intermittently coupled to deployment specific reward evaluation, particle reweighting, resampling and perturbation. 

\subsection{Local Refinement}
\label{sec:local_refinement}
Toward the end of the diffusion trajectory, $\forall t \in \mathcal{T}_{LR}$, our reward alignment method is complemented by a gradient-free local refinement of near feasible grasps ($\mathcal{T}_{LR}=\{8,9\}$ for single arm and 
$\{223,224,225\}$ for dual arm). For each particle, a local neighborhood is constructed by applying translational and rotational perturbations along the $\mathfrak{se}(3)$ tangent space. Candidate joint configurations are first obtained through a damped Jacobian update for efficient screening through local rewards, after which the best candidate is subjected to hard feasibility checks as the global reward. The candidate Jacobian update $\Delta q^a$ is given by
\begin{equation}
    \Delta q^a = J^\top
    \left(
        JJ^\top+\eta^2 I_6
    \right)^{-1}
        \begin{bmatrix}
        \mathbf{t}_{\mathrm{c}}-\mathbf{t}(q^a)\\
        \Logmap\!\left(\mathbf{R}_{\mathrm{c}}\mathbf{R}(q^a)^\top\right)
    \end{bmatrix}
    \label{eq:damped_jacobian}
\end{equation}
where $H_{\mathrm{c}}=(\mathbf{R}_{\mathrm{c}},\mathbf{t}_{\mathrm{c}})$ and $H(q^a)=(\mathbf{R}(q^a),\mathbf{t}(q^a))$ denote the perturbed and parent grasp poses respectively, $J$ is the
end-effector spatial Jacobian evaluated at $q^a$, and $\eta>0$ is the damping coefficient. The local objective focuses on fine grained gripper-object interactions while whole arm constraints are enforced by rejecting candidates that violate or degrade the parent's feasibility constraints. The local and global steering objectives are instantiated from the same deployment-specific feasibility measures, defined next.

\subsection{Grasp Feasibility Rewards}\label{subsec:feasibility_rewards}
We evaluate the Tweedie estimate of each denoised grasp using geometric and kinematic quantities induced by the embodiment $\mathcal{R}$ and the deployment environment $\mathcal{E}$. The evaluators are queries directly at the inference time and need not be differentiable with respect to the grasp pose, unlike gradient-based approaches \cite{arm_aware,ggp}. For each arm $a$, the candidate grasp pose is first resolved to a joint configuration $q^a$ through multi-start inverse kinematics using \texttt{PyTorch Kinematics} \cite{pytorch_kinematics}.

\begin{figure*}[!t]
    \centering
    \includegraphics[width=\textwidth]{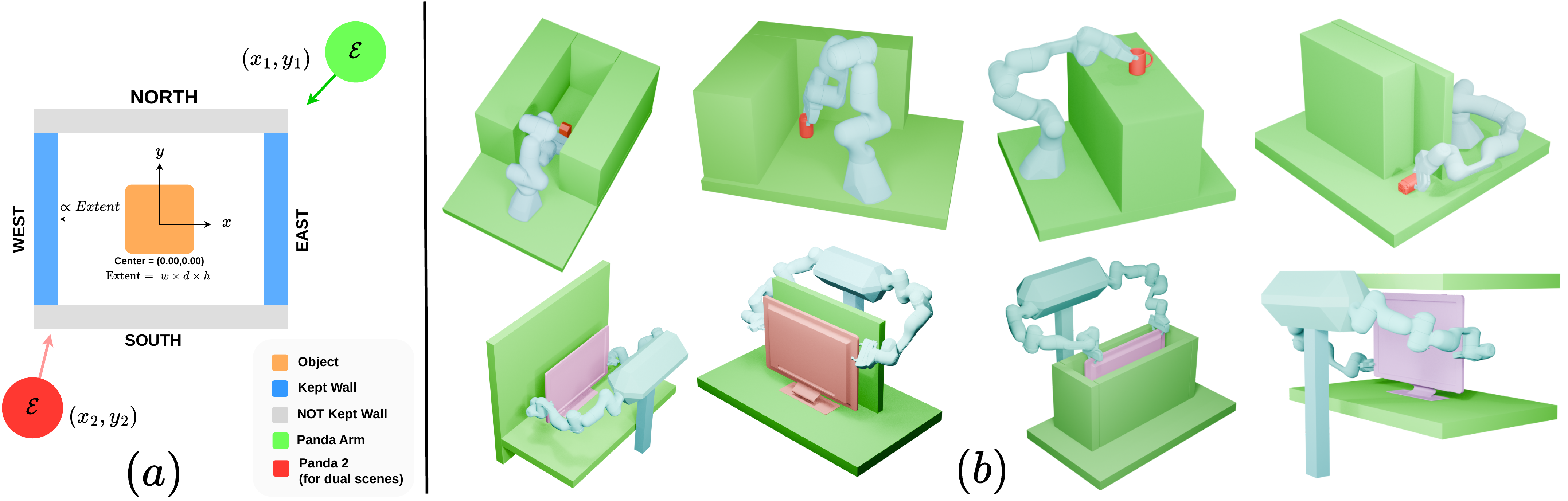}
    \captionsetup{font=footnotesize}
    \caption{\textbf{Evaluation settings.}
       \textbf{(a) Large Scale Evaluation.} Top-down schematic of the scene generation scheme (see legend); the local frame at the object's center shows its position and extent $(w\times d \times h)$, while $\theta$ and (x, y) at the sampled Panda arm(s) base give its heading and location relative to the world origin. The wall labels (NORTH / SOUTH / EAST / WEST) indicate the four candidate wall positions, of which kept vs. not-kept walls are distinguished per scene. \textbf{(b) Curated Scenes.} Top-Row: \textbf{Single Arm Scenes} from L-R: S1. \textit{Corridor}, S2. \textit{Room Corner}, S3. \textit{Shelf Top} and S4. \textit{Partition Board}. Bottom-Row: \textbf{Dual Arm Scenes} from L-R: S1. \textit{Table} S2. \textit{Wall} S3. \textit{Box} S4. \textit{Shelf} }
    \label{fig:curated_scenes_large_scale}
    \vspace{-20pt}
\end{figure*}

\paragraph{\textbf{Arm Collision Avoidance}} We represent the robot links using collision spheres for efficient geometric collision evaluation. For an arm $a$, arm-environment clearance is evaluated against the environment obstacles using their signed distance functions,
\begin{equation}
    d_{\mathrm{env}}^a=\min_{i,o}
    \left[
        \operatorname{SDF}_{o}
        \!\left(c_i^a(q^a)\right)-\rho_i^a
    \right]
    \label{eq:env_clearance}
\end{equation}
where $c_i^a(q^a)$ and $\rho_i^a$ denote the center and radius of the $i$-th collision sphere, and $o \in \mathcal{E}$ indexes the environment obstacles. In contrast, arm-object clearance is evaluated directly from the point cloud $P$ using sphere-point signed distances,
\begin{subequations}
\begin{align}
    d_{\mathrm{obj},ij}^{a} & = \left\|c_i^a(q^a)-p_j\right\|_2-\rho_i^a,
    \qquad p_j\in P
    \label{eq:obj_pairwise_clearance}\\
    d_{\mathrm{obj}}^a &
    =
    \operatorname{BottomKMean}_{k}
    \left(
        \left\{
            d_{\mathrm{obj},ij}^{a}
        \right\}_{i,j}
    \right)
    \label{eq:object_clearance}
\end{align}
\end{subequations}
where $\operatorname{BottomKMean}_{k}$ denotes the mean distances between the $k$ smallest pairwise sphere point distances, excluding the gripper collision spheres. This clearance is then used for the gripper-object term in the local refinement objective to rank candidate grasp perturbations. 

Self collision $d_{self}^a$ is evaluated from the minimum pairwise signed clearance among spheres of the same arm, whereas for dual-arm grasps the inter-arm clearance $d_{inter}$ is evaluated between the two collision sphere sets in the shared world frame. The pairwise sphere clearance is
\begin{subequations}
    \begin{align}
        d_{inter,ij} &
        =
        \|c_i^L(q^L)-c_j^R(q^R)\|_2
        -\rho_i^L-\rho_j^R
        \label{eq:inter_arm_distance}\\
        d_{\mathrm{inter}} & =\min_{i,j}\{d_{inter,ij}\}
    \end{align}
\end{subequations}
To deploy with dense geometry (Real-World in Sec.~\ref{subsec:evaluation_settings}), we instantiate the environment distance queries via the Euclidean Signed Distance Fields (ESDFs) using \texttt{cuRoboV2} \cite{curobov2}. 

\paragraph{\textbf{Grasp Reachability}}
\label{para:grasp_reachability}
For each arm $a$, the candidate grasp is mapped to a joint configuration $q^a$ through multi-start IK, selecting the solution with minimum weighted pose residual $
    e_{\mathrm{IK}}^a
    =
    w_{\mathbf{t}} e_{\mathrm{pos}}^a
    +
    w_{\mathbf{R}} e_{\mathrm{rot}}^a
    \label{eq:ik_residual}
$, where $e_{\mathrm{pos}}^a=\|\mathbf{t}^a-\mathbf{t}_{\mathrm{FK}}^a(q^a)\|_2$ and
$e_{\mathrm{rot}}^a=\|\Logmap(\mathbf{R}^a\mathbf{R}_{\mathrm{FK}}^a(q^a)^\top)\|_2$
measure translational and rotational errors between the candidate grasp pose
$H^a$ and its forward-kinematic realization
$H_{\mathrm{FK}}^a(q^a)$, respectively. This residual provides the kinematic reachability
measure, with smaller values indicating closer realization of the desired
grasp pose.

\paragraph{\textbf{Reward scaling and aggregation}}The resulting signed clearances and kinematic residuals are converted to bounded rewards using a $\tanh(.)$ function $\mathcal{S}(d;s)=\tanh\!\left(d/s \right)$.
where $s$ is a fixed characteristic scale. Thus all clearance measures $d \in \{d^a_{env},d^a_{obj}, d^a_{self}, d^a_{inter}\}$ are mapped as $\mathbf{r}=\mathcal{S}(d;s)$, while the non-negative IK residual is negated before normalization as $
    \mathbf{r}_{\mathrm{IK}}^a
     =
     \tanh\!\left(
        -e_{\mathrm{IK}}^a/
         s_{\mathrm{IK}}
     \right)$

For a single-arm grasp, the deployment reward is given by
\vspace{-2mm}
\begin{equation}
    \mathbf{r}_{single} = 
    w_{\mathrm{IK}}\mathbf{r}_{\mathrm{IK}}
    +w_{\mathrm{env}}\mathbf{r}_{\mathrm{env}}
    +w_{\mathrm{obj}}\mathbf{r}_{\mathrm{obj}}
    +w_{\mathrm{self}}\mathbf{r}_{\mathrm{self}}
    \label{eq:single_reward}
\end{equation}
\vspace{-3mm}

For a dual-arm grasp, the two predicted grasp poses are first assigned to the
physical arms by selecting the assignment with the lowest total IK residual. Assuming $\mathbf{r}^a_{\mathrm{single}}$ for the reward in
Eq. \ref{eq:single_reward} evaluated on each arm $a\in\{L,R\}$, the dual-arm reward is
\vspace{-1mm}
\begin{equation}
    \mathbf{r}_{\mathrm{dual}} = w_{\mathrm{inter}}\mathbf{r}_{\mathrm{inter}} +
    \min_{a\in\{L,R\}} \mathbf{r}^a_{\mathrm{single}}
    \label{eq:dual_reward}
\end{equation}
where $\mathbf{r}_{\mathrm{inter}}$ penalizes collisions between the two arms.
Taking the minimum over arms ensures that a grasp is rewarded only if it is
feasible for both.

\section{Experiments}

\subsection{Baselines}
We compare our method with the following baselines: 
\paragraph{\textbf{Plain Sampling}} Generate $N$ grasps per object using the base diffusion sampler and then evaluate them all. 

\paragraph{\textbf{Rejection Filter}} Generate $N$ grasps per object using the base diffusion sampler and filter out grasps that don't have strictly positive scalar reward (Sec.~\ref{subsec:feasibility_rewards}). This can generate less than $N$ grasps per object on average. 

\paragraph{\textbf{Oversample ($\bm{M} \rightarrow \bm{N}$)}} Generate $M$ grasps where $M>N$, and retain only top $N$ by reward.  (M, N) is (500,100) for dual-arm and (150, 50) for single-arm.

\begin{table*}[!t]
    \centering
    \renewcommand{\arraystretch}{1.12}
    \setlength{\tabcolsep}{4.5pt}
    \begingroup
    \small
    \newcommand{\basec}{{\scriptsize\,(base)}}
    \newcommand{\dlt}[1]{\ensuremath{{+}#1}}
    \newcommand{\dltn}[1]{\ensuremath{{-}#1}}

    \begin{tabular}{c lcc|cc|c|cc|cc}
        \toprule
        &
        \textbf{Method}
            & \textbf{Reach.}$\uparrow$ & \textbf{Coll.}$\downarrow$
            & \multicolumn{2}{c|}{\textbf{CS} $\uparrow$}
            & \textbf{SR}$_{hold}$ $\uparrow$
            & \multicolumn{2}{c|}{\textbf{{SR}$_{hold}$ +CS} $\uparrow$}
            & \textbf{Time}$\downarrow$ & \textbf{VRAM}$\downarrow$ \\

        \cmidrule(lr){5-6}
        \cmidrule(lr){8-9}

        &
            & \% & \%
            & @10 & @all
            & \%
            & @10 & @all
            & s\,($+$base) & GB\,($+$base) \\

        \midrule

        % ============================================================
        % Single-Arm
        % ============================================================

        \multirow{6}{*}{\rotatebox[origin=c]{0}{\textit{Single-Arm}}}
        & Plain sampling
            & 72.51 & 76.24
            & 40.90 & 15.60
            & 18.44
            & 24.00 & 8.93
            & 0.34\basec
            & 0.53\basec \\

        & Rejection filter\dag
            & 89.50 & 71.41
            & 47.14 & 18.81
            & 22.31
            & 27.02 & 10.78
            & \dlt{1.12}
            & \dlt{0.00} \\

        & Oversampling
            & 97.84 & 56.88
            & 60.34 & 41.06
            & 29.77
            & 34.84
            & 22.80
            & \dlt{1.52}
            & \dlt{0.08} \\

        & Gradient guidance
            & 76.41
            & 31.52
            & 69.64
            & 57.71
            & 21.51
            & 32.01
            & 18.52
            & \dlt{11.23}
            & \dlt{0.01} \\

        \cmidrule(l){2-11}

        & \textbf{Ours, O}
            & \cellcolor{second}98.26
            & \cellcolor{second}28.41
            & \cellcolor{second}72.99
            & \cellcolor{second}70.42
            & \cellcolor{second}38.84
            & \cellcolor{second}37.84
            & \cellcolor{second}38.13
            & \dlt{3.32}
            & \dlt{0.00} \\

        & \textbf{Ours, O+LR}
            & \cellcolor{best}98.43
            & \cellcolor{best}26.56
            & \cellcolor{best}74.54
            & \cellcolor{best}72.41
            & \cellcolor{best}41.41
            & \cellcolor{best}40.31
            & \cellcolor{best}40.79
            & \dlt{3.72}
            & \dlt{0.31} \\

        \midrule

        % ============================================================
        % Dual-Arm
        % ============================================================

        \multirow{6}{*}{\rotatebox[origin=c]{0}{\textit{Dual-Arm}}}
        & Plain sampling
            & 40.91 & 88.85
            & 19.55 & 4.88
            & 67.52
            & 17.17 & 4.21
            & 10.46\basec
            & 1.28\basec \\

        & Rejection filter\dag
            & 76.22 & 86.30
            & 20.11 & 9.26
            & 78.05
            & 17.48 & 8.23
            & \dltn{0.16}
            & \dlt{0.00} \\

        & Oversampling
            & 78.70 & 82.99
            & 33.92 & 14.55
            & 79.91
            & 30.30 & 12.74
            & \dlt{14.67}
            & \dlt{1.86} \\

        & Gradient Guidance
            & 61.74 & 75.51
            & 49.07 & 15.97
            & 67.55
            & 43.43 & 12.48
            & \dlt{175.36}
            & \dlt{1.56} \\

        \cmidrule(l){2-11}

        & \textbf{Ours, O}
            & \cellcolor{second}98.57
            & \cellcolor{second}49.23
            & \cellcolor{best}62.18
            & \cellcolor{second}50.77
            & \cellcolor{best}85.11
            & \cellcolor{best}54.98
            & \cellcolor{second}45.03
            & \dlt{2.43}
            & \dlt{0.00} \\

        & \textbf{Ours, O+LR}
            & \cellcolor{best}98.80
            & \cellcolor{best}46.47
            & \cellcolor{second}61.93
            & \cellcolor{best}53.53
            & \cellcolor{second}84.64
            & \cellcolor{second}54.28
            & \cellcolor{best}47.59
            & \dlt{6.47}
            & \dlt{0.09} \\

        \bottomrule
    \end{tabular}

    \endgroup

    \captionsetup{font=footnotesize}
    \caption{
        \textbf{Large-scale scene evaluation.}
        We generate $N$ grasps for all methods; metrics are averaged across all scenes.
        \textbf{Reach.}~is the reachability rate
        (IK convergence; for dual-arm, both arms must converge).
        \textbf{Coll.}~refers to the collision rate.
        \textbf{CS@k}~is constraint satisfaction among the top-$k$
        ranked grasps per scene.
        \textbf{SR}$_{hold}$~denotes stability under all pull axes.
        \textbf{SR}$_{hold}$\textbf{+CS@k}~is the joint fraction that is stable,
        reachable, and collision-free.
        \textbf{Time}~and \textbf{VRAM}~are end-to-end generation latency (s) and
        peak GPU memory (GB), reported as the change relative to plain sampling of
        the same setting; the plain-sampling row gives the absolute baseline.
        Rows below the rule are ours.
        \colorbox{best}{Green}/\colorbox{second}{Orange}
        indicate best/second-best per column within each setting.
        \dag~Rejection filter on average generates fewer than $N$ grasps per scene.
        Our method is denoted by shorthand \textbf{O} and with local refinement added to it, shorthand is \textbf{O+LR}.
    }
    \vspace{-20pt}
    \label{tab:large_scale}
\end{table*}

\paragraph{\textbf{Gradient Guidance}}
Additionally, we compare with an existing gradient guidance method~\cite{arm_aware}. We extend the single arm implementation to the dual-arm case by introducing an additional inter-arm collision cost term $f_{inter}$ computed using the differentiable log-sum-exp smoothing from Appendix~A.2 of~\cite{embodisteer}. We additionally augment per-arm objective with an arm--object clearance soft constraint, using the sphere--point signed distance in Eq.~\ref{eq:obj_pairwise_clearance} and the same smooth SoftPlus collision cost as in~\cite{arm_aware}. For each Tweedie estimated grasp pose $(\hat{H}_0^L,\hat{H}_0^R)$ and feasible IK solution set $\mathcal{Q}^L$ and $\mathcal{Q}^R$, we evaluate the original single-arm cost $f^L$ and $f^R$. We jointly select the optimal pair $(q^{L,\star},q^{R,\star})$, where $q^{L,\star}\in\mathcal{Q}^L$ and $q^{R,\star}\in\mathcal{Q}^R$, by solving
\vspace{-1mm}
\begin{equation}
\begin{aligned}
(q^{L,\star},q^{R,\star})
&=
\arg\min_{q^L,q^R}
\Big[
f^L(\hat{H}_0^L,q^L)
+
f^R(\hat{H}_0^R,q^R) \\
&\qquad\qquad
+
\lambda_{inter}f_{inter}(q^L,q^R)
\Big]
\end{aligned}
\label{eq:dual_arm_cost}
\end{equation}
\vspace{-3mm}

where $\lambda_{inter}$ is the Lagrange multiplier for the inter-arm collision constraint, while the per-arm multipliers are contained in $f^L$ and $f^R$. The task space guidance gradients are then evaluated at the current grasp poses' Tweedie estimate $(\hat{H}_0^L,\hat{H}_0^R)$ and the jointly selected IK solutions $(q^{L,\star},q^{R,\star})$,
\begin{equation}
\begin{aligned}
\nabla_{\hat{H}_0^L}F
&=
\nabla_{\hat{H}_0^L}f^L
+
\lambda_{inter}(J_L^+)^{\mathsf T}\nabla_{q^{L,\star}}f_{inter},\\
\nabla_{\hat{H}_0^R}F
&=
\nabla_{\hat{H}_0^R}f^R
+
\lambda_{inter}(J_R^+)^{\mathsf T}\nabla_{q^{R,\star}}f_{inter},
\end{aligned}
\label{eq:dual_arm_gradients}
\end{equation}

where $J_L^+$ and $J_R^+$ denote the damped pseudoinverses of the left and right arm Jacobians, respectively. These gradients are concatenated to form a joint 12D guidance update for the given grasp pair.

\subsection{Evaluation settings}
\label{subsec:evaluation_settings}
\textbf{Large-scale evaluation.} Following the scheme in Fig.~\ref{fig:curated_scenes_large_scale}(a), we randomly spawn up to four walls around an object and randomize the base position of each Franka Emika Panda arm, retaining 50 scenes per object after discarding those with no executable grasp. Using 18 single-arm objects from \cite{acronym} and 12 dual-arm objects from \cite{dg16m}, with $N{=}50$ grasps and $N{=}100$ grasp pairs per scene respectively, yielding 45,000 single-arm grasps and 60,000 dual-arm grasp pairs, all evaluated in \texttt{MuJoCo}.

\textbf{Curated scenes.} For both single and dual arm settings we use
four representative scenes, each with a single object in a distinct
configuration (Fig.~\ref{fig:curated_scenes_large_scale}(b)). We evaluate all
methods on two embodiments, a UR5e and a Franka Emika Panda, both with Robotiq
grippers, using the execution metrics of Sec.~\ref{subsec:metrics} and $N{=}50$
grasps or grasp pairs per scene.

\textbf{Real world.} To demonstrate sim-to-real transfer, we evaluate on xArm7
manipulators with xArm grippers across three scenes per object, using two objects for each of the single and dual arm settings (Fig.~\ref{fig:object-results-wide}). The \textit{Easy} scene tests reachability and ground collision avoidance, with additional inter-arm clearance in the dual arm setting. The \textit{Medium} scene constrains the workspace with a corridor-like configuration, and the \textit{Hard} scene combines narrow clearances with restricted approach directions. Point clouds are captured from two RealSense views Fig.~\ref{fig:object-results-wide} and fused to compute the ESDF. Grasps are executed using the \texttt{cuRoboV2} motion planner.

\subsection{Metrics}
\label{subsec:metrics}
Unless otherwise specified, all metrics reported are calculated for \(N\) total grasp candidates generated. 

\textbf{Reachability (Reach).}
A grasp is considered \textit{reachable} if an IK solution exists with a positional and rotational error less than 1 cm and 2 degrees respectively (Sec.~\ref{subsec:feasibility_rewards} (b)). For dual-arm grasp, both arms must admit valid IK solutions. 

\textbf{Collision (Coll).} Each grasp is instantiated via its corresponding joint configuration produced by the method and the arm is checked for collisions with the environment, with itself, or in the dual-arm case, with each other.

\textbf{Constraint Satisfaction (CS@\(k\)).}
A grasp is constraint-satisfying if it is reachable and collision-free.
CS@\(\{10,\mathrm{all}\}\) reports the fraction of the top-\(k\) grasps that
satisfy this, where grasps are ranked by the harmonic mean of the normalized values of the model specific grasp quality scores (force closure scores for DAGDiff, and discriminator confidence for GraspGen). For plain sampling, rejection filtering, oversampling, and our method, the feasibility score is Eq.~\ref{eq:single_reward} (single arm) or Eq.~\ref{eq:dual_reward}
(dual arm); for Gradient Guidance, we use the objective of~\cite{arm_aware} (or
Eq.~\ref{eq:dual_arm_cost} for our dual-arm extension), min-max normalized
over all sampled grasps in the scene. 

\textbf{Holding Success Rate (SR\(_{\mathrm{hold}}\)).}
Each grasp is instantiated in \texttt{MuJoCo} with its arm
joint configuration, and the object but without environment obstacles.
Following~\cite{ggp}, a \(3\,\mathrm{N}\) pull is applied sequentially along the \(\pm x\), \(\pm y\),
and \(\pm z\) axes of the object frame, and the grasp succeeds if the object's
center of mass stays within \(5\,\mathrm{cm}\) of its initial position.

\textbf{SR\(_{\mathrm{hold}}\)+CS@\(k\).}
Under the same harmonic ranking, the fraction of top-\(k\) grasps which satisfy the constraints and pass SR\(_{\mathrm{hold}}\). This is what we define as \textit{feasibility}.

\textbf{Task Success Rate (SR\(_{\mathrm{task}}\)).}
The fraction of grasps that a motion planner in \texttt{MuJoCo} can execute
through the full task trajectory (e.g., lifting, pulling the object out) in
the presence of environment obstacles while maintaining a stable grasp.

\subsection{Results}

\begin{table}[!t]
    \centering
    \renewcommand{\arraystretch}{1.12}
    \setlength{\tabcolsep}{2.0pt}
    \begingroup
    \footnotesize

    \resizebox{\columnwidth}{!}{%
    \begin{tabular}{cl|cccc|cccc}
        \toprule
        \multirow{2}{*}{\textbf{Arm}}
        & \multirow{2}{*}{\textbf{Method}}
        & \multicolumn{4}{c|}{\textbf{Single-Arm} (SR$_{\mathrm{task}}$)}
        & \multicolumn{4}{c}{\textbf{Dual-Arm} (SR$_{\mathrm{task}}$)} \\

        \cmidrule(lr){3-6}
        \cmidrule(lr){7-10}

        &
        & \textbf{S1} & \textbf{S2} & \textbf{S3} & \textbf{S4}
        & \textbf{S1} & \textbf{S2} & \textbf{S3} & \textbf{S4} \\

        \midrule

        % ============================================================
        % Franka
        % Single: S1 Corridor, S2 Room Corner, S3 Shelf Top, S4 Partition Board
        % Dual:   S1 Table,    S2 Wall,        S3 Box,       S4 Shelf
        % ============================================================

        \multirow{6}{*}{\rotatebox[origin=c]{90}{\textit{Franka}}}
        & Plain sampling
        & 10/50 & 3/50 & 4/50 & 22/50
        & 7/50 & 7/50 & 0/50 & 3/50 \\

        & Rejection filter
        & 10/29 & 3/34 & 4/27 & 22/37
        & 7/26 & 7/16 & 0/14 & 3/9 \\

        & Oversampling
        & \cellcolor{second}48/50 & 13/50 & 29/50 & \cellcolor{second}49/50
        & 39/50 & 45/50 & 1/50 & 9/50 \\

        & Gradient Guidance
        & 24/50 & 14/50 & 37/50 & 34/50
        & 30/50 & 39/50 & \cellcolor{second}18/50 & \cellcolor{second}29/50 \\

        & \textbf{Ours, O}
        & \cellcolor{best}50/50 & \cellcolor{second}33/50 & \cellcolor{second}46/50 & \cellcolor{best}50/50
        & \cellcolor{second}48/50 & \cellcolor{second}48/50 & \cellcolor{best}35/50 & \cellcolor{best}50/50 \\

        & \textbf{Ours, O + LR}
        & \cellcolor{best}50/50 & \cellcolor{best}36/50 & \cellcolor{best}48/50 & \cellcolor{best}50/50
        & \cellcolor{best}50/50 & \cellcolor{best}49/50 & \cellcolor{best}35/50 & \cellcolor{best}50/50 \\

        \midrule

        % ============================================================
        % UR5e
        % ============================================================

        \multirow{6}{*}{\rotatebox[origin=c]{90}{\textit{UR5e}}}
        & Plain sampling
        & 11/50 & 6/50 & 1/50 & 21/50
        & 9/50 & 11/50 & 3/50 & 0/50 \\

        & Rejection filter
        & 11/34 & 6/29 & 1/25 & 21/45
        & 7/26 & 5/9 & 3/22 & 1/7 \\

        & Oversampling
        & \cellcolor{best}47/50 & 23/50 & 10/50 & \cellcolor{best}50/50
        & 37/50 & 35/50 & 34/50 & 6/50 \\

        & Gradient Guidance
        & 39/50 & 13/50 & 31/50 & \cellcolor{second}27/50
        & 48/50 & 42/50 & \cellcolor{second}37/50 & \cellcolor{second}33/50 \\

        & \textbf{Ours, O}
        & \cellcolor{second}46/50 & \cellcolor{second}35/50 & \cellcolor{second}44/50 & \cellcolor{best}50/50
        & \cellcolor{second}49/50 & \cellcolor{second}48/50 & \cellcolor{best}50/50 & \cellcolor{best}50/50 \\

        & \textbf{Ours, O + LR}
        & \cellcolor{best}47/50 & \cellcolor{best}37/50 & \cellcolor{best}46/50 & \cellcolor{best}50/50
        & \cellcolor{best}50/50 & \cellcolor{best}50/50 & \cellcolor{best}50/50 & \cellcolor{best}50/50 \\

        \bottomrule
    \end{tabular}%
    }

    \captionsetup{font=footnotesize}
    \caption{%
        \textbf{Curated-scene evaluation.}
        \textbf{SR}$_{\mathrm{task}}$ (successful/total grasps) on curated scenes (refer Fig. \ref{fig:curated_scenes_large_scale}) for single- and dual-arm grasps across arm types.
        \textbf{SR}$_{\mathrm{task}}$ is the fraction of generated grasps that are constraint-satisfying and successfully transport the object in \texttt{MuJoCo}.
        \colorbox{best}{Green}/\colorbox{second}{Orange} indicate best/second-best per column within each arm. Our method is denoted by shorthand \textbf{O} and with local refinement added to it, shorthand is \textbf{O+LR}.
    }
    \vspace{-21pt}
    \label{tab:curated_scene_results}
    \endgroup
\end{table}

\subsubsection{\textbf{Quantitative Evaluation}} \label{subsec:quantitative_eval}

We report large scale results in Tab.~\ref{tab:large_scale} and curated scene results in Tab.~\ref{tab:curated_scene_results}. 
Plain sampling represents the unguided prior and, unless mentioned otherwise, serves as the reference for all comparisons, since the complete set of executable grasps in a scene is unknown.

\textit{Sample Efficiency:}
Rejection filter produces only 35.77 single- and 37.3 dual-arm grasps on average in large scale evaluation, reflecting less sample efficiency than other methods that retain all single and dual grasps respectively. Unlike oversampling, which improves feasibility by discarding grasps based on reward, our method reallocates particle mass toward feasible modes, increasing yield of feasible grasps for a fixed number of samples.

\textit{Constraint Satisfaction:}
Only $15.6\%$ of single-arm grasps and $4.9\%$ of dual-arm grasps from the prior are reachable and collision-free. Our method (\textbf{O}) raises constraint satisfaction over the full population(\textbf{CS}@$\mathrm{all}$) by $+55$pp and $+46$pp, with near-complete reachability in both settings and a collision rate reduced by $48$ pp and $40$ pp. Meanwhile gradient guidance trails our method (\textbf{O}) by $13$ pp in the single-arm case, but on moving to dual-arm it lags by $35$pp, since resampling reallocates the population to \textit{\textbf{feasible modes}} regardless of their distance in $SE(3)\times SE(3)$. These modes exist in the prior but carry little mass, so drawing more samples, as in oversampling, recovers them only in proportion to that mass, adding $+26$ pp and $+10$ pp, whereas our method (\textbf{O}) concentrates the population on them at a fixed budget.

\textit{Stability:}
Steering toward feasibility does not cost grasp stability. $\mathbf{SR}_{\mathrm{hold}}$ for our method (\textbf{O}) over all generated grasps improves by $+20$ pp (single-arm) and $+18$ pp (dual-arm) over plain sampling, and by a similar margin over gradient guidance.
Gradient guidance adds a reward gradient to every reverse step, while our method \textbf{(O)} leaves the transition kernel untouched, and every surviving trajectory remains close to a sample path of \textit{\textbf{the learned prior}}. Our method \textbf{(O)} improves feasibility, ($\mathbf{SR}_{\mathrm{hold}}+\mathbf{CS}$) over all grasps, by $+29$ pp and $+41$ pp over plain sampling and by $+20$ pp and $+32$ pp over gradient guidance. The feasible grasps of the baselines are confined to a few top-ranked samples while ours span the population, emphasizing our methods ability to provide population-wide correction. 

\textit{Effect of Local Refinement:}
Local refinement acts only on near-feasible particles, so it leaves the top-ranked grasps untouched and instead attempts to make marginal ones feasible. The result is a consistent gain in collision-free grasps and population-wide feasibility of $2$--$3$ pp in both settings, with stability and ranking unchanged or slightly improved. Since it accounts for the entire memory overhead ($0.3$\,GB) and most of the dual-arm latency ($4$ of $6.5$\,s), we report \textbf{(O+LR)} as an option rather than as part of the core steering. 

\begin{figure}[t]
    \centering
    \includegraphics[width=1\columnwidth]{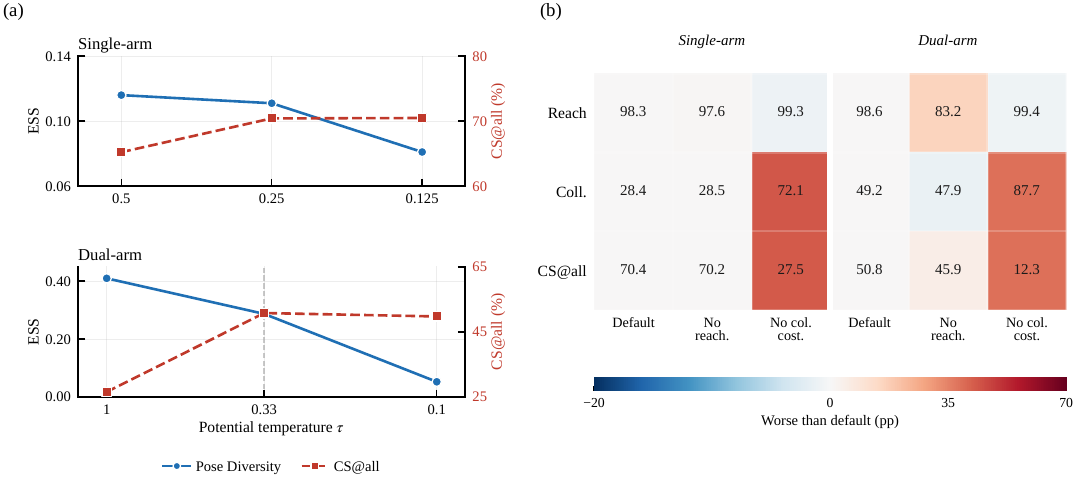}
    \captionsetup{font=footnotesize}
    \caption{\textbf{Ablations}: a) Effect of Temperature $\tau$ on Effective Sample Size and Constraint Satisfaction. b) Effect of rewards on Reachability and Collision 
    }
    \vspace{-21pt}
    \label{fig:ablation}
\end{figure}

\textit{Downstream execution and cross-embodiment transfer:}
Tab.~\ref{tab:curated_scene_results} evaluates $\mathbf{SR}_{\mathrm{task}}$, i.e.\ whether generated grasps can actually be executed by a motion planner in \texttt{MuJoCo} in the presence of the obstacles, on the curated scenes of Fig.~\ref{fig:curated_scenes_large_scale}. Across the 16 scene embodiment combinations, our method \textbf{(O)} is best or tied-best in 7 scenes and, with local refinement \textbf{(O+LR)}, in all 16. The gains are largest where the prior fails (Ex. In the dual-arm \textit{Box} scene (S3) on the Franka, plain sampling yields no executable grasp pair and $5\times$ oversampling a single pair), and gradient guidance recovers about a third, and our method (\textbf{O}) more than two thirds ($35/50$), showing that the improved static feasibility translates to reliable downstream manipulation execution.

\textit{Real World:}
After three trials per scene, our method (\textbf{O+LR}) achieves \textbf{30/36} successful executions (Table~\ref{tab:real_success}). Failures mainly occur for the bucket in the \textit{Hard} setting due to asymmetric loading and the lack of compliant torque control.

\textit{Computational cost:}
Steering adds $3.3$\,s to single-arm and $2.4$\,s to dual-arm generation, against $11.2$\,s and $175.4$\,s for gradient guidance, which evaluates IK and its gradient at every diffusion step, in contrast to our intermittent steering \textbf{(O)}. The cost of steering is roughly constant, while the cost of oversampling grows with the prior and the sample factor it requires. For the cheap single-arm prior, $3\times$ oversampling is about half our overhead, but for dual-arm grasping where $5\times$ oversampling is needed, our method is $6\times$ faster and adds no GPU memory against nearly $2$\,GB overhead. Using local (\textbf{O+LR}) adds a mere 0.4s and 4s overhead in single and dual settings to our method respectively, while improving collision avoidance and reachability, and hence the overall constraint satisfaction over our method \textbf{(O)}.

\subsubsection{\textbf{Ablation Study}}
\label{subsec:ablation}

All ablations use our method \textbf{(O)} to isolate the  hyperparameters. We first vary the temperature $\tau$ (Fig.~\ref{fig:ablation}a). Lowering $\tau$ raises constraint satisfaction until it saturates, at $\tau\approx0.25$ for single-arm and $\tau\approx0.33$ for dual-arm, while diversity keeps falling as the population collapses onto a few high-reward modes.
Fig.~\ref{fig:ablation}b removes individual reward terms. Without the collision reward the collision rate returns to nearly the level of plain sampling in both settings, so reachability alone steers particles toward the robot but not around obstacles. 
Removing the reachability reward has a much smaller effect, since the collision reward already disfavors unreachable poses which tend to place the arm in collision. This coupling is strongest for single-arm grasping, where objects are small and an unreachable pose usually forces the arm into collision. For the larger dual-arm objects the coupling is weaker, and reachability drops by $15$ pp, highlighting the necessity of the rewards. More results are provided in the video.

\captionsetup[subfigure]{
    skip=1pt,
    font=footnotesize
}

\newcommand{\resultimg}[1]{%
    \includegraphics[
        width=\linewidth,
        height=0.18\columnwidth,
        keepaspectratio
    ]{#1}%
}

\begin{figure}[t]
    \centering

    \subcaptionbox{Cylinder\label{fig:cylinder}}[0.245\columnwidth]{%
        \centering
        \resultimg{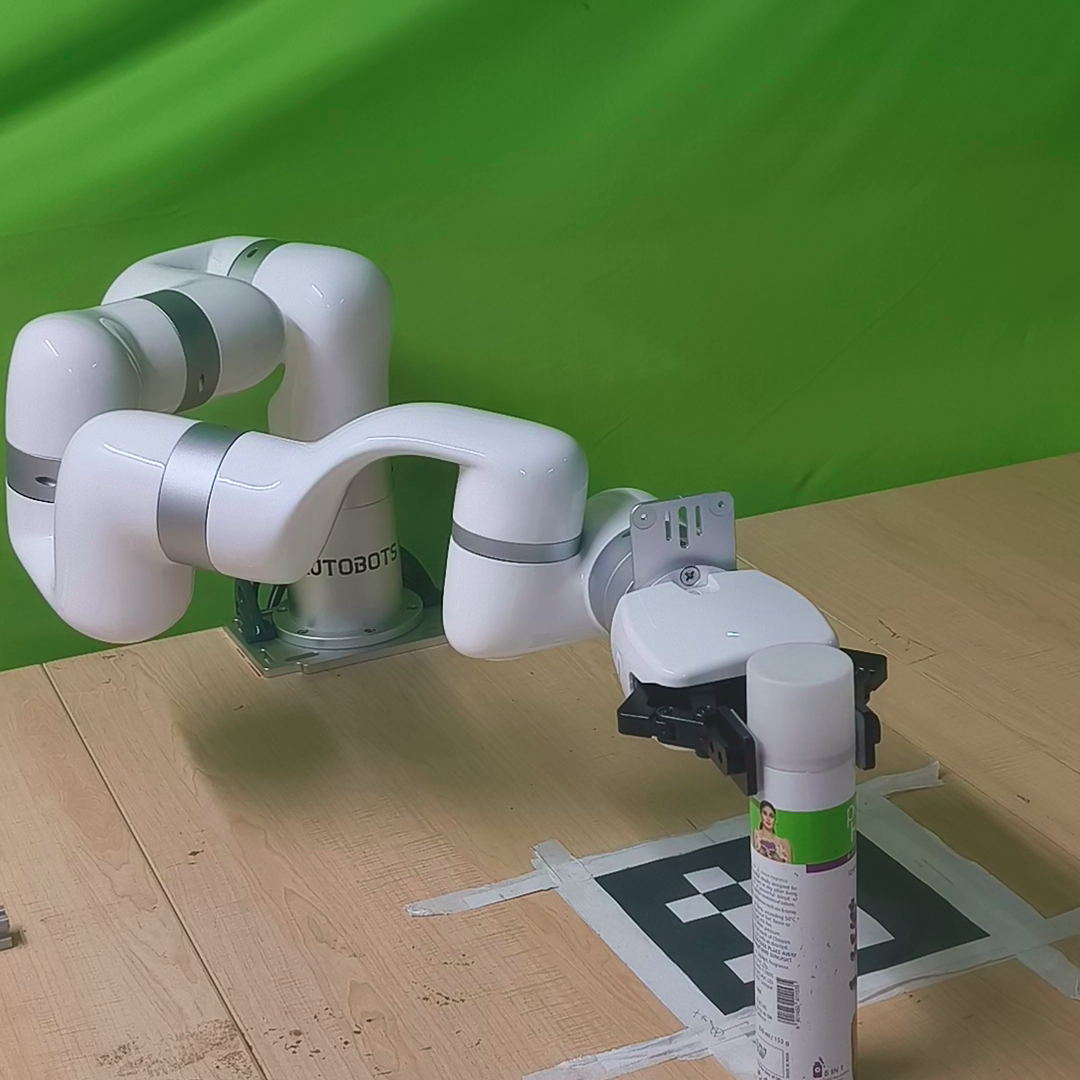}\\[-0.8mm]
        \resultimg{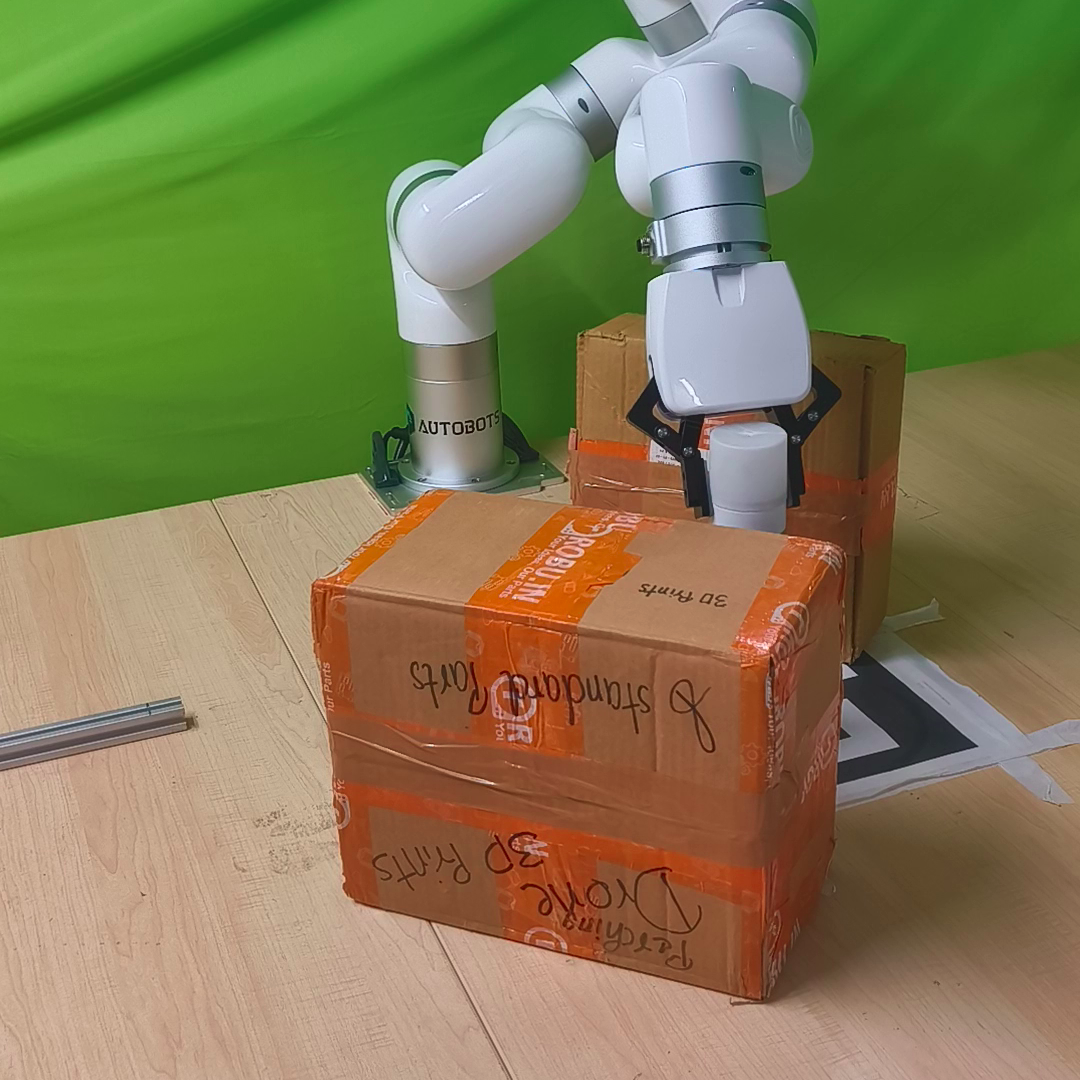}\\[-0.8mm]
        \resultimg{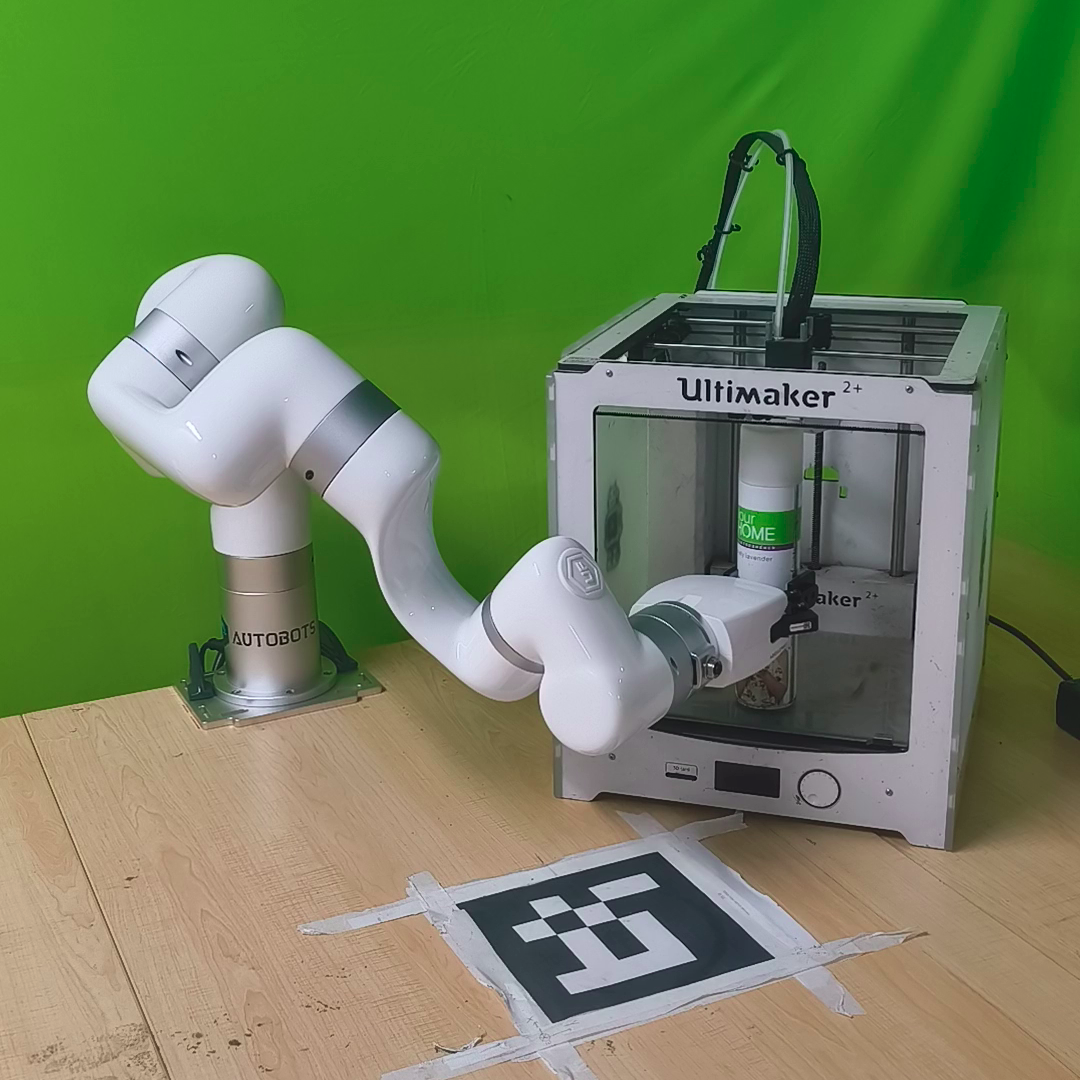}%
    }%
    \hspace{-0.5mm}%
    \subcaptionbox{Bucket\label{fig:bucket}}[0.245\columnwidth]{%
        \centering
        \resultimg{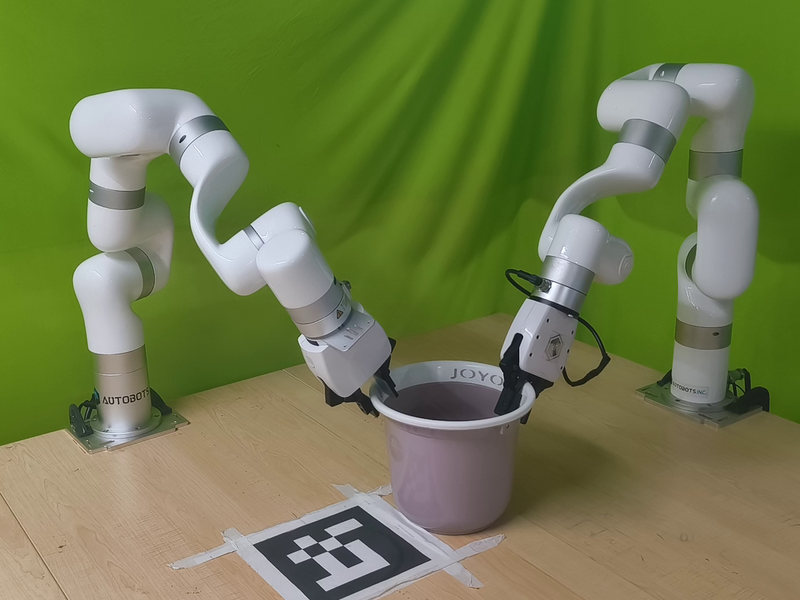}\\[-0.8mm]
        \resultimg{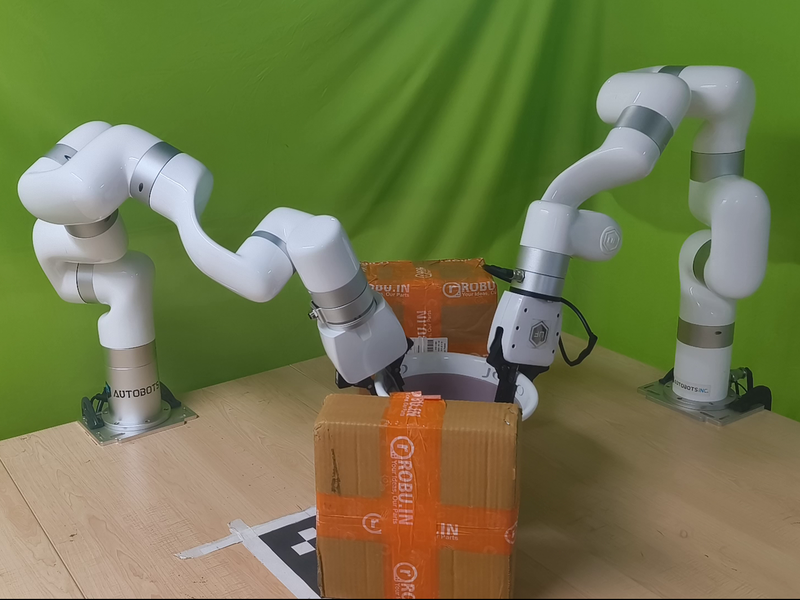}\\[-0.8mm]
        \resultimg{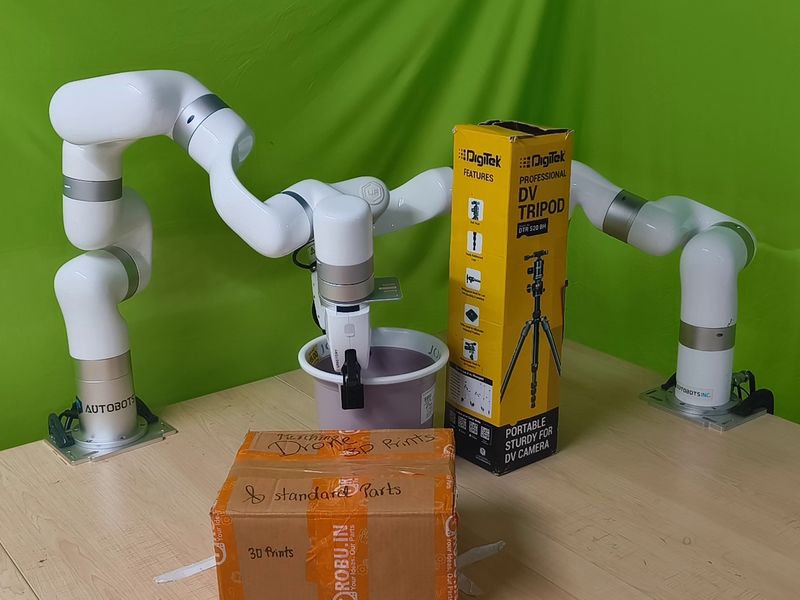}%
    }%
    \hspace{-0.5mm}%
    \subcaptionbox{Bowl\label{fig:bowl}}[0.245\columnwidth]{%
        \centering
        \resultimg{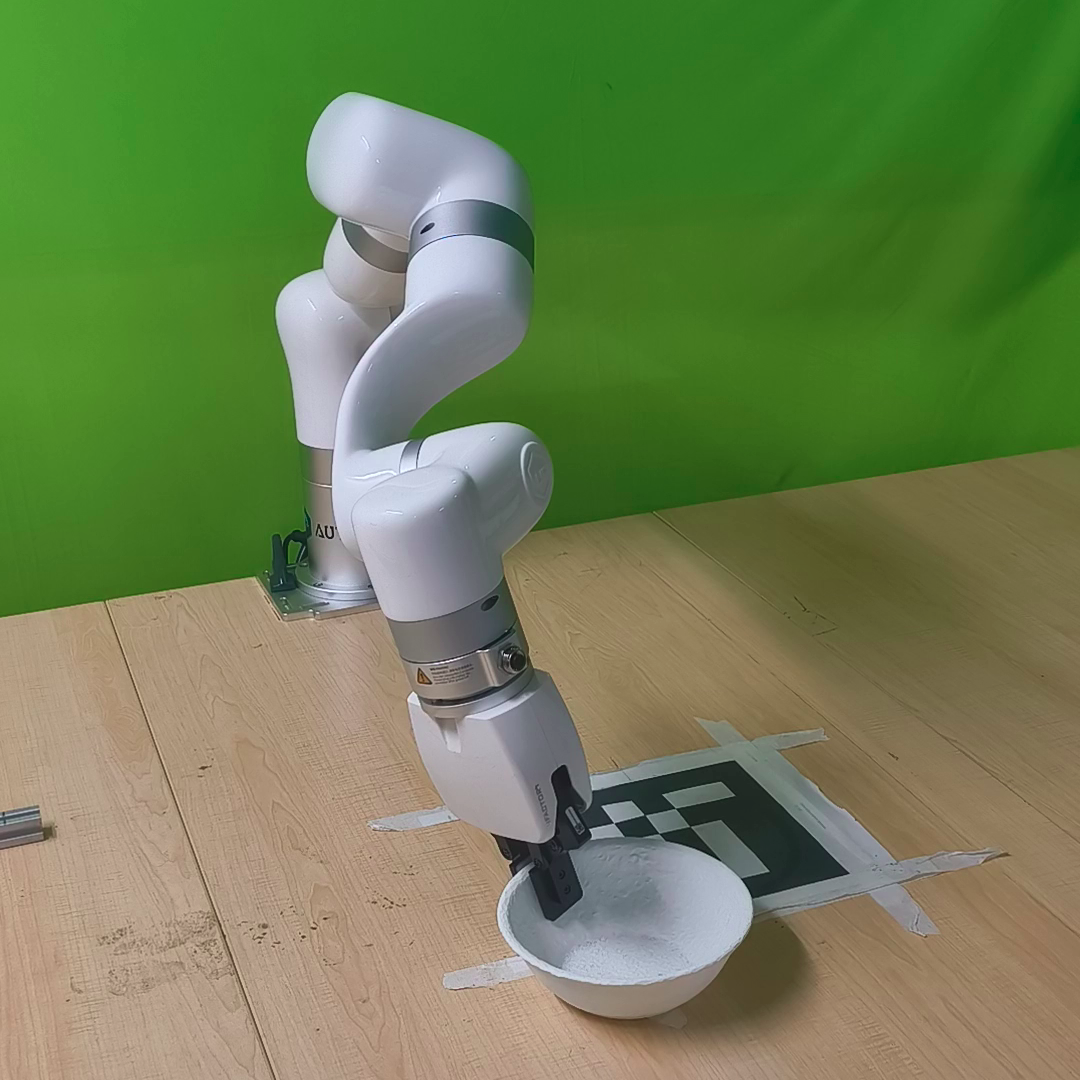}\\[-0.8mm]
        \resultimg{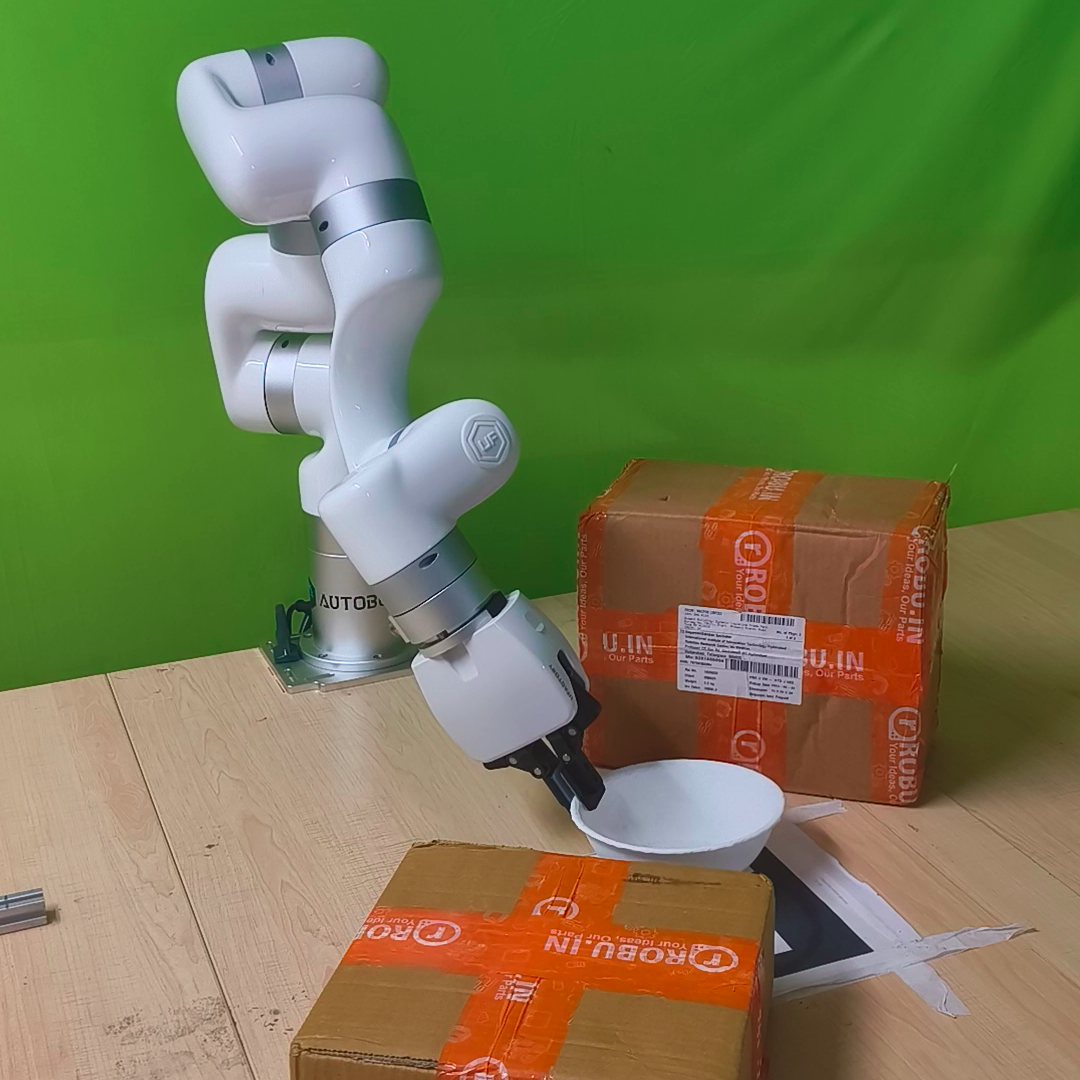}\\[-0.8mm]
        \resultimg{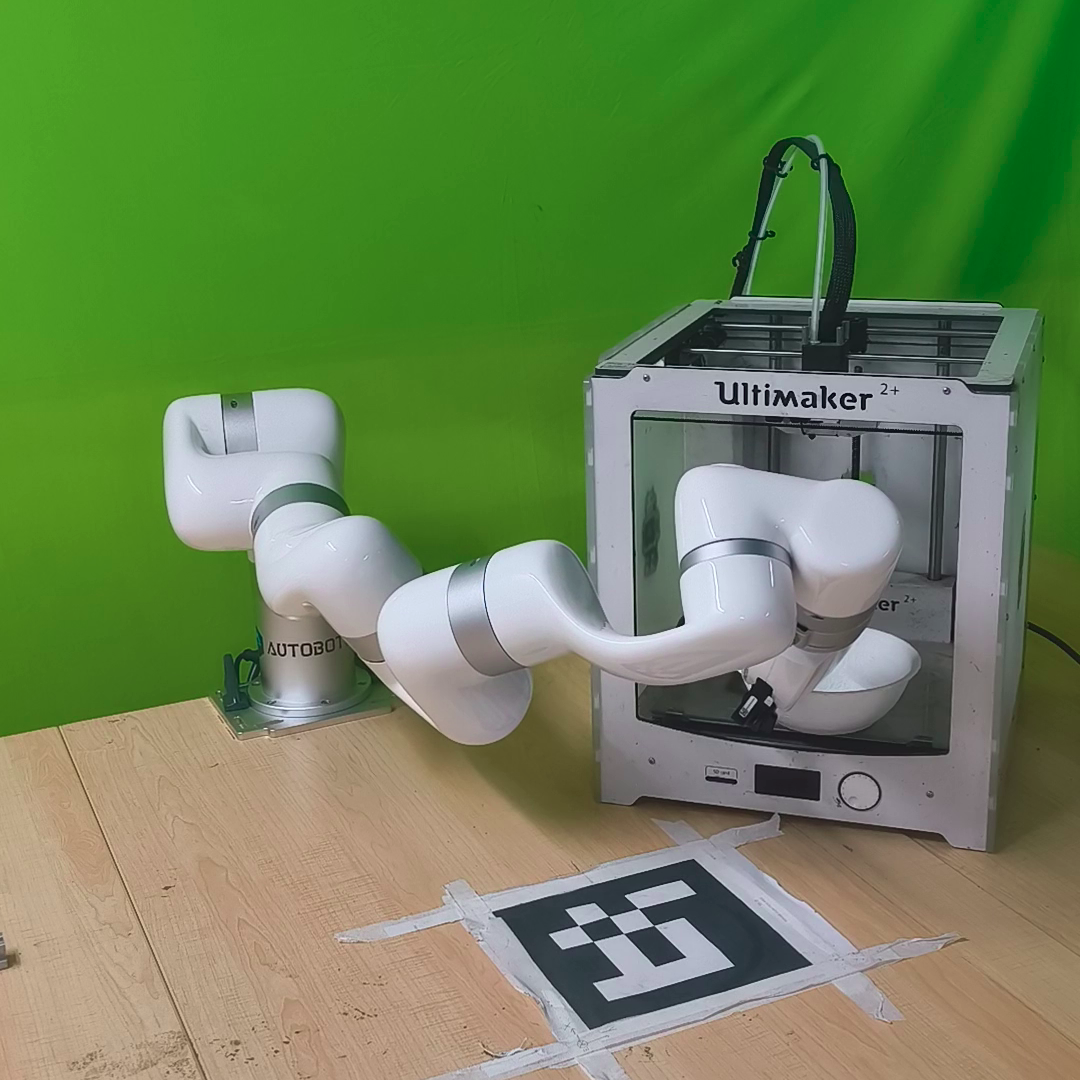}%
    }%
    \hspace{-0.5mm}%
    \subcaptionbox{Tray\label{fig:tray}}[0.245\columnwidth]{%
        \centering
        \resultimg{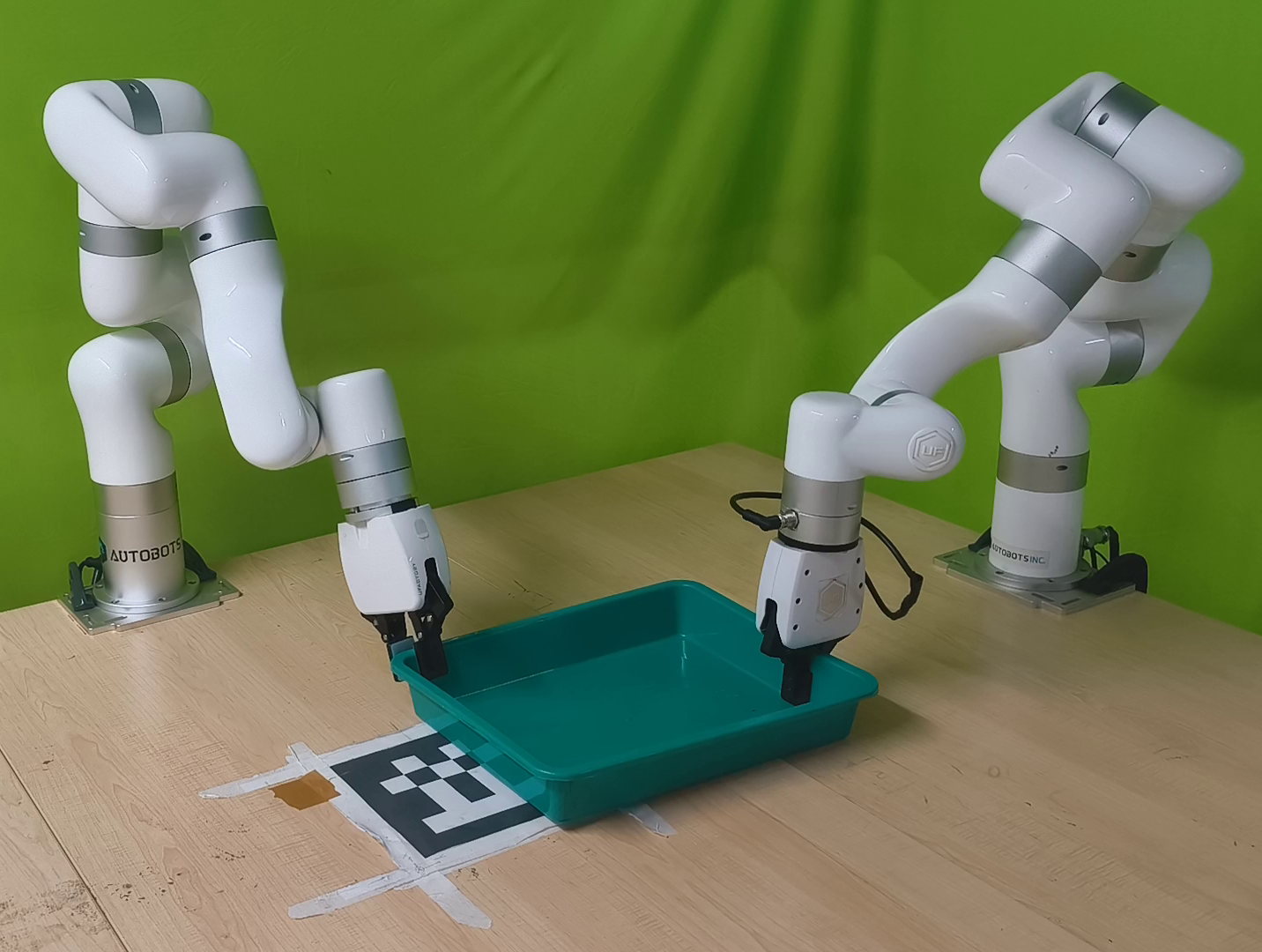}\\[-0.8mm]
        \resultimg{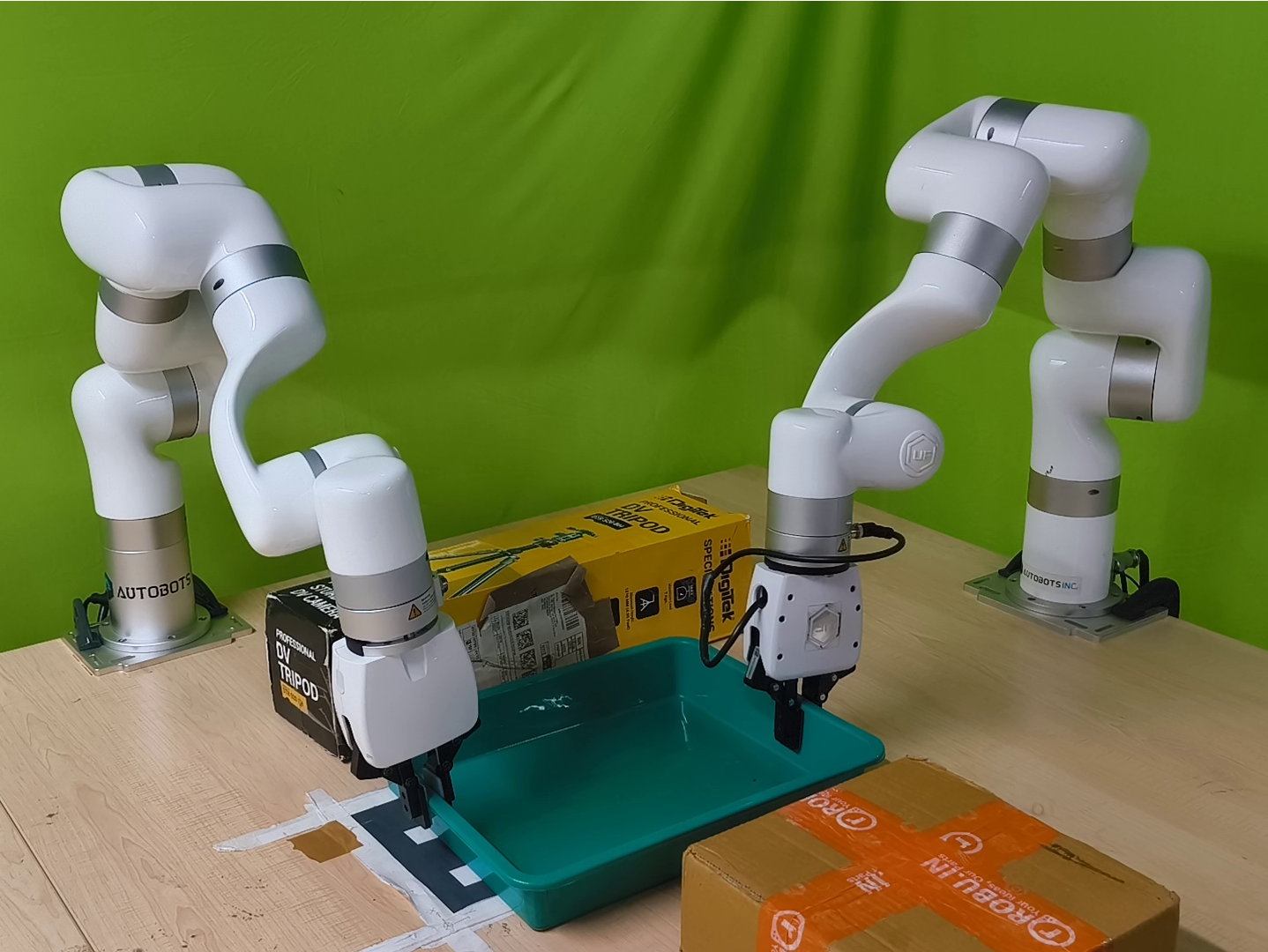}\\[-0.8mm]
        \resultimg{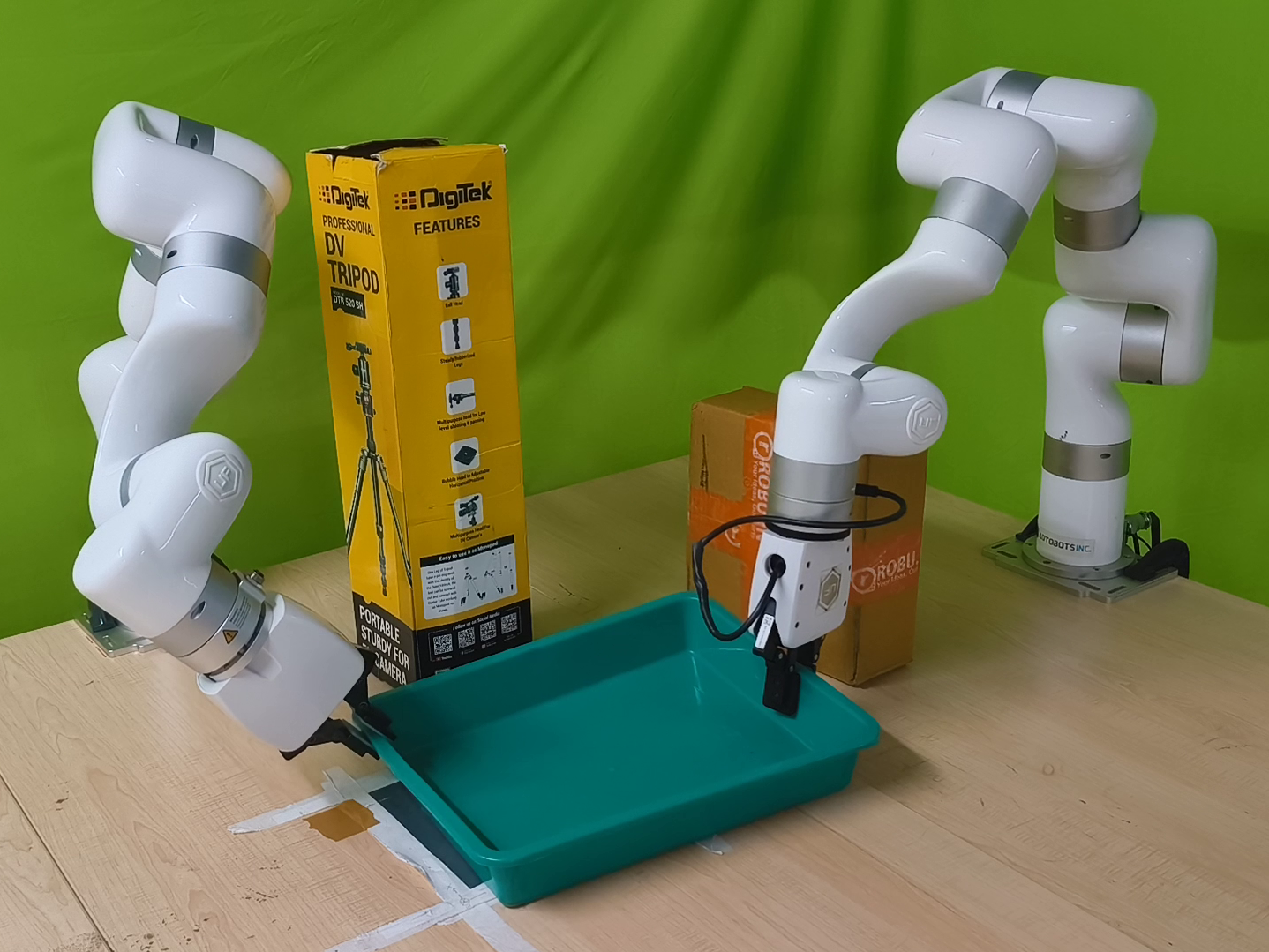}%
    }
    \captionsetup{font=footnotesize}
    \caption{\textbf{Real-world grasping.}
    Columns correspond to the four objects; rows from top to bottom
    correspond to \textit{Easy}, \textit{Medium}, and \textit{Hard} scenarios.}
    \vspace{-5pt}
    \label{fig:object-results-wide}
\end{figure}

\begin{table}[t]
    \centering
    \renewcommand{\arraystretch}{1.0}
    \setlength{\tabcolsep}{3.0pt}
    \begin{tabular}{l l c c c c}
        \toprule
        \textbf{Arm Type} & \textbf{Object} & Easy & Medium & Hard & \textbf{Total}\\
        \midrule
        \multirow{2}{*}{Single} & Bowl     & 3/3 & 3/3 & 3/3 & \textbf{9/9}\\
        & Cylinder & 3/3 & 3/3 & 2/3 & \textbf{8/9}\\
        \midrule
        \multirow{2}{*}{Dual}  & Bucket   & 3/3 & 2/3 & 1/3 & \textbf{6/9}\\
        & Tray     & 3/3 & 2/3 & 2/3 & \textbf{7/9}\\
        \bottomrule
    \end{tabular}
    \captionsetup{font=footnotesize}
    \caption{\textbf{Real-world success}. 9 trials per object across varying difficulty.}
    \vspace{-20pt}
    \label{tab:real_success}
\end{table}

\section{Conclusion}
\label{sec:conclusion}
We presented \textbf{\coolname}, a training-free framework that introduces embodiment awareness into frozen grasp diffusion models through population-level inference-time steering. By reweighting and resampling particles using kinematic and collision rewards, the method enables population reallocation to feasible modes while preserving the closeness to the underlying grasp prior. Experiments across single- and dual-arm models, multiple embodiments, constrained scenes, and hardware show improved feasible and stable grasp generation over plain sampling and local gradient guidance at substantially lower cost than aggressive oversampling. 
As future work, we plan to extend the framework to incorporate sparse or long-horizon objectives, such as manipulability, future trajectory feasibility, and contact dynamics, through trajectory-level or dynamics-aware guidance.

\bibliographystyle{IEEEtran}

\bibliography{bibtex}

\end{document}